\pdfoutput=1
\documentclass[conference]{IEEEtran}
\IEEEoverridecommandlockouts
\usepackage{amsmath,amssymb,amsfonts}
\usepackage{algorithmic}
\usepackage{graphicx}
\usepackage{float}
\usepackage{rotating}
\usepackage{threeparttable}
\usepackage{booktabs}
\usepackage{array}
\usepackage{textcomp}
\usepackage{xcolor}
\usepackage{cite}

\newcolumntype{L}[1]{>{\raggedright\let\newline\\\arraybackslash\hspace{0pt}}m{#1}}
\newcolumntype{C}[1]{>{\centering\let\newline\\\arraybackslash\hspace{0pt}}m{#1}}
\newcolumntype{R}[1]{>{\raggedleft\let\newline\\\arraybackslash\hspace{0pt}}m{#1}}

\begin{document}

\title{Identifying patterns of proprioception and target matching acuity in healthy humans\\
\thanks{This work has been submitted to the IEEE for possible publication. Copyright may be transferred without notice, after which this version may no longer be accessible.}
}

\author{\IEEEauthorblockN{Jacob Carducci}
\IEEEauthorblockA{\textit{Mechanical Engineering} \\
\textit{Johns Hopkins University}\\
Baltimore, MD, USA \\
jcarduc1@jh.edu}
\and
\IEEEauthorblockN{Jeremy D. Brown}
\IEEEauthorblockA{\textit{Mechanical Engineering} \\
\textit{Johns Hopkins University}\\
Baltimore, MD, USA \\
jdelainebrown@jhu.edu}

}

\maketitle

\begin{abstract}
Traditional approaches to measurement in upper-limb therapy have gaps that electronic sensing and recording can help fill. We highlight shortcomings in current kinematic recording devices, and we introduce a wrist sensing device that performs multimodal sensing during single-axis rotation. Our goal is to characterize normative kinesthetic perception and real-world performance as a multimodal sensory ``fingerprint'' that can serve as a reference point for identifying deficit in persons affected by stroke, and then as a jumping point for later neuroscientific interrogation. We present an experiment involving psychophysical measurements of passive stimuli discrimination, matching adjustment acuity, and ADL performance in 11 neurologically-intact persons. We found that passive velocity sense and active position sense of healthy controls, measured by velocity discrimination and position matching respectively, correlated in rank with each other, but other score comparisons of acuity or task performance had no statistically significant correlations. We also found that participants differed in acuity between passive and active velocity sense, which supports current understanding about muscle spindle activation being modulated by conscious motor command. The potential for our null correlation results to reveal dissociable aspects of deficit is discussed, as well as implications for future neuroscientific study with more kinematic measures and larger datasets.
\end{abstract}

\begin{IEEEkeywords}
rehabilitation, neuroscience, robotics, proprioception, wrist, kinematics, psychophysics
\end{IEEEkeywords}

\section{Introduction}
Impairment of sensation and/or motion in the wrist has a profound effect on a person’s ability to interact with their environment. Sensory impairment receives less research attention than motor impairment, but 85\% of acute stroke reports highlight some form of sensory deficit overall \cite{Kim1996DiscriminativeStroke} and 40\% in the upper limbs \cite{Rathore2002CharacterizationSymptoms}. Sensory impairment can also negatively impact recovery rates of motor function \cite{Bolognini2016TheRehabilitation} and quality of life \cite{Sullivan2008SensoryIntervention}. Despite functional tests that characterize limitation in general upper-limb sensorimotor ability, these assessments tend to rely on  discrete qualitative observation of therapists, or use measurements from physical tools like stopwatches, goniometers, and rulers that are cumbersome to capture and transcribe into datasets \cite{Lang2013AssessmentMaking}. Test-retest reliability in clinical settings can also be an issue, especially for tactile tests \cite{Fonseca2018FunctionalProperties,Connell2012MeasuresReview}, position gauge tests \cite{Pilbeam2018TestretestSetting,Connell2012MeasuresReview}, and sensory components of functional tests \cite{Lin2004PsychometricPatients}. Therefore, there is a pressing need to offer more reliable and repeatable sensory assessments to identify sensory deficits. 

Somatosensory information, including limb position and force, coexists and can integrate in complex ways to support cognitive and motor functions \cite{Azanon2016MultimodalRepresentation}. Severity of brain disability can impair sensory modalities differently and impact networks of intermodal influence \cite{Bolognini2016TheRehabilitation}. These patterns of relationships among modalities can be represented as multimodal profiles of somatosensory signatures or ``fingerprints.'' This modeling philosophy of categorical featuring and comparison has been applied in previous stroke \cite{Senadheera2023ProfilingLimb} and developmental disability \cite{Little2018SensoryDevelopment} studies. In the stroke study, in particular, different clusters of impairment profiles in touch and position sense were identified, suggesting shared neural correlates in the parietal operculum that influence object recognition \cite{Senadheera2023ProfilingLimb}. However, proprioception beyond position and shape sense was not investigated. Without a comprehensive assessment of sensory ability across all sensory modalities and stages of deficit, it can be difficult to precisely determine progress made by the patient and intervention dosage needed to address the whole profile of disease.

Increased efforts have been made recently to identify and train sensory ability in addition to motor function, and to specifically restore sensory ability from impairment  \cite{Arya2022ActiveReview}. For example, a randomized controlled trial, termed the Study of the Effectiveness of Neurorehabilitation on Sensation (SENSe), evaluated active sensory therapy of 50 stroke patients with somatosensory deficit. Discrimination training of these patients in a cross-over study design did result in aggregate sensory improvement in tactile and position-sense tasks, which was maintained at 6 and 24 weeks after intervention \cite{Carey2011SENSe:Sensation}.  More generally, there can be a correlation between improvements in motor function and matching acuity during recovery \cite{Cherpin2019APatients}. Similarly, proprioceptive deficit has been treated in recent interventions where baseline and improvement was distinct from cognitive or motor scores \cite{Young2022WristStroke,Mochizuki2019MovementExoskeleton}. In other words, treating sensory and motor deficits separately may or may not influence the other type’s recovery depending on patient and context, but both should be treated nevertheless. The key to treating the two types of deficit separately and collectively are richer assessment approaches that are better informed by sound neuroscience.

While motor and somatosensory functions are complementary in manipulation and cannot be easily untangled, some components can be identified and treated separately given the biomechanical structure of the human arm. It is well-established that proprioceptive and kinesthetic sensation operates through different receptors and neural pathways parallel to afferent motor activation \cite{Proske2012TheForce,Proske2018KinestheticSenses,Prochazka2021Proprioception:Neurophysiology}. Muscles spindles parallel to muscle contractile units, receptor endings in the joints, and golgi tendon organs all provide a neurophysiological basis as distinct proprioceptors that influence different sensations and perception of muscle-space position, velocity, and force. Several recent studies \cite{Li2016CorrelationMethods, Nagai2016ConsciousSubmodality, Marini2017TheProprioception, Niespodzinski2018RelationshipProprioception} and a comprehensive metastudy \cite{Horvath2023TheReview} have supported this notion of independent kinesthetic senses through test measures of stimuli discrimination and matching. Sensation of limb position may additionally depend on the relationship between the human and the environment. Internal limb comparisons and external pointing tasks can result in different performance \cite{Chen2021PositionAngles,Proske2021TwoProprioception}, suggesting that integrating vision and internal body schema may add additional context to the perception of limb position. Even direction and magnitude of target stimulus can play a role in positional matching \cite{DAntonio2021RoboticApproach}. Therefore, it is possible that distinct sensory deficits may manifest differently depending on the specific proprioceptive affect and context, creating a multidimensional ``fingerprint'' of sensory ability impacted by disease.

Different joints of the human arm have different contributions to active kinesthesia sense, especially the wrist \cite{Walsh2013TheWrist,Dounskaia1998HierarchicalPatterns}. The wrist is important and unique in upper-limb manipulation during activities of daily living (ADL) \cite{Anderton2022MovementLiving,Franko2008FunctionalMotion}, therefore it is important to identify and treat any deficit in the wrist to improve upper-limb function. You can train the wrist holistically with the whole arm in multi-joint treatment \cite{Xu2023Multi-jointPoststroke}, given that deficits can rank correlate (but vary) across joints \cite{Xu2023Multi-jointPoststroke,AbiChebel2022JointPerception} and that the central nervous system (CNS) optimizes endpoint position more than joint angles \cite{Fuentes2010WhereTasks}. Alternatively, you can train specifically at the joint level. Our focus for this engineering and assessment feasibility study is joint-specific and wrist-specific for two reasons primarily: (1) the wrist is uniquely important for movement correction \cite{Dounskaia1998HierarchicalPatterns} and tends to have more severe and more variable deficit compared to other joints \cite{Xu2023Multi-jointPoststroke,AbiChebel2022JointPerception}; and (2) 1-DOF tasks help reduce confounding variables during human subjects research, and these tasks can later build up to multi-joint study. There have been efforts to leverage sensory electronics to record continuous quantitative measures of generalized kinematics and kinetics \cite{Koeppel2020Test-RetestRehabilitation}, and use these signals to quantify psychophysical metrics in the whole arm \cite{Contu2017ProprioceptiveManipulator,Deblock-Bellamy2018QuantificationDisplay} and in the wrist \cite{Rinderknecht2016ReliableParadigm,Cappello2014EvaluationDevice}. However, current normative datasets of psychophysical measures at the wrist are just emerging for position-sense \cite{Cappello2015Robot-AidedProprioception,Cappello2015WristAssessment,Contu2015PreliminaryWrist,Marini2016Robot-AidedWorkspace} and torque-sense \cite{Feyzabadi2013HumanDesign,Vicentini2010EvaluationSystem}. For persons affected by neurological disease, available wrist psychometric data is especially sparse, under-powered, and only available for position-sense \cite{Elangovan2019ASurvivors, Contu2018WristPatients}. Despite reliable assessment of velocity discrimination existing for the ankle \cite{Westlake2007VelocityAdults} and assessment of velocity reproduction at the knee \cite{Nagai2016ConsciousSubmodality}, no velocity matching or discrimination data has been established for the wrist in either healthy or patient populations. 

It has been argued that robotics and wearables with rich sensing can inform better rehabilitation models that incorporate plasticity of sensory improvement over time \cite{Reinkensmeyer2016ComputationalRecovery}. If we could determine measures of discrimination and acuity quickly and precisely, even for patients with limited motion, we can build large sets of normative and wrist-impaired data with multiple dimensions of psychophysical and kinematic markers to allow for complex disease categorization and individualized treatment. Therefore, our engineering goal is to develop an assessment device and associated protocol that is capable of capturing signals from the wrist via multimodal sensing in a compact package that can scale up datasets of discrimination and acuity quickly. Furthermore, a related goal is to demonstrate this device's ability to test neuroscientific hypotheses and verify predictions about somatosensory patterns. After device construction, we conducted a preliminary battery of assessments on neurologically-intact participants to: 1) evaluate the device's assessment capabilities; and 2) determine any preliminary relationship between measures of ability that have the potential to reveal an impairment cluster pattern of association in patient populations. Specifically, we investigated contralateral target pointing via a gauge matching test, ipsilateral discrimination and matching via robotic display of kinematic cues (i.e.,~position, velocity, and torque), and functional ability via virtual tasks of kettle pouring and door manipulation.

Overall, we hypothesized that extents of somatosensory and functional ability can correlate with each other, meaning that sensing poorly in one aspect can predict poor sense or function in another modality. From this, we predicted that at least one correlation will exist across handedness laterality, cognitive ability, functional ability, psychometric acuity, and ADL performance at the wrist. The reason for demonstrating correlations between scores was to see how many scoring components are needed at minimum to acquire a useful sensory ``fingerprint''; if two scores are highly correlated in rank across a population, then only one of those scores is needed as a ``fingerprint'' component to explain ability. That way, less tests are needed to generate approximately the same ``fingerprint'' for describing ability and deficit, so it becomes faster and easier to diagnose and treat. To verify the prediction of any co-varied association, rank correlation analysis (e.g.,~Spearman's coefficient calculation) was conducted.

While our first prediction/hypothesis may indicate a monotonic correlation between good passive sense and good active sense for perceiving wrist angle, it does not specify if these sensory capabilities are distinct (and separable) aspects of proprioception. Therefore, we additionally hypothesized that kinesthetic senses can differ in acuity depending on task context. More specifically, sensing a rotational wrist speed passively can be more or less sensitive and accurate compared to sensing speed while dynamically activating wrist muscles via conscious motor command. Furthermore, using the same hand to accurately identify position can differ from replicating a wrist position using the other hand. Accordingly, we predicted that contralateral position matching, active ipsilateral position matching, and passive ipsilateral position discrimination will have distinctly different angle acuity scores (at the wrist).  We also predicted that passive and active velocity senses at the wrist will be distinctly different in speed acuity scores. To detect differences in modality scores between task contexts, we used a statistical inference test of ranking (e.g.,~a Friedman test).  

\section{Methods}
\subsection{Equipment Setup}
\begin{figure}[tb] 
   \centering
   \includegraphics[width=0.95\columnwidth]{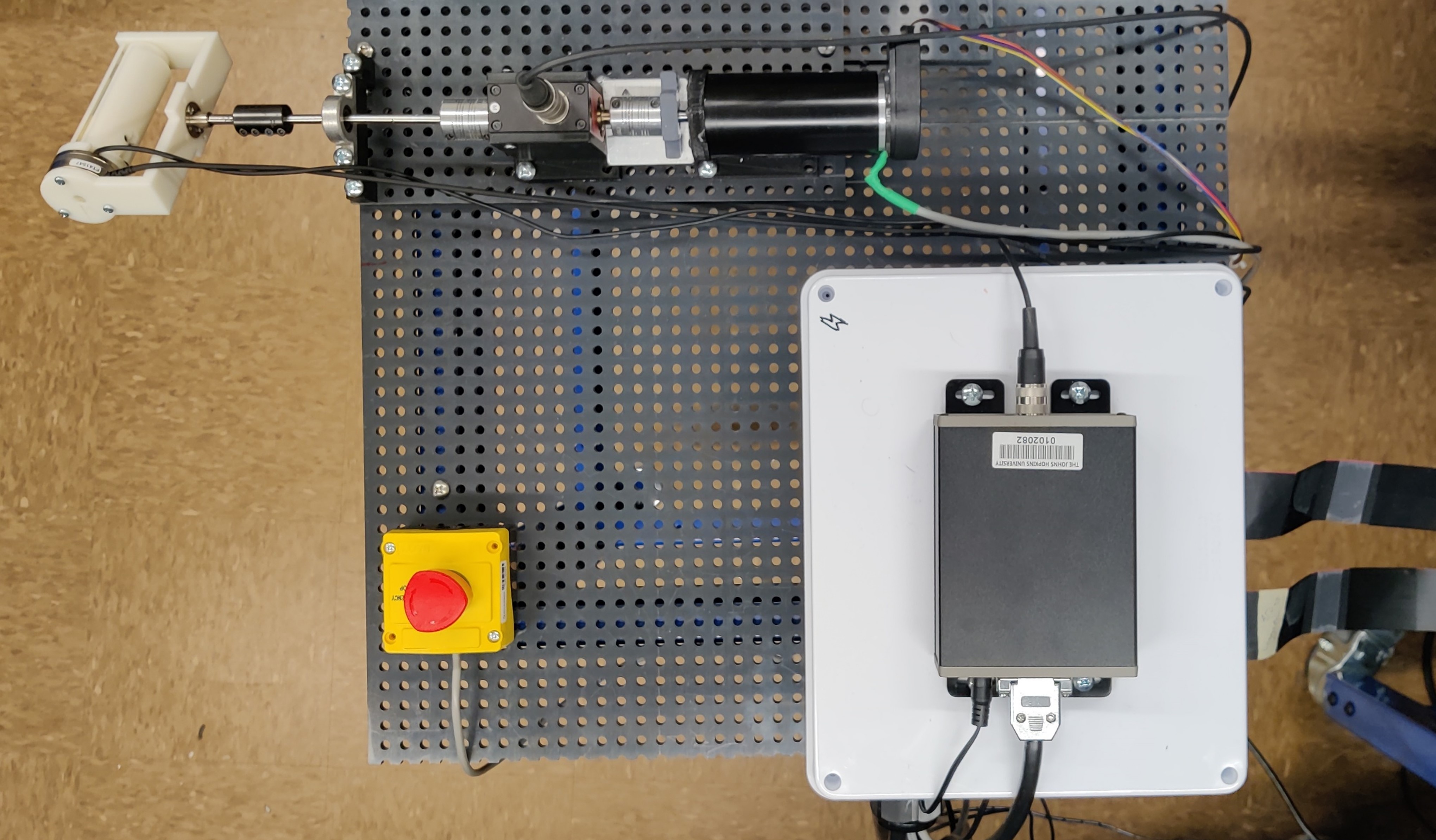}
   \caption{Birds-eye view of the WRIST testbed, with the motor-driven wrist interface at top and the signal conditioning and amplification electronics in the white box at the bottom.}
   \label{fig:wristsetup}
\end{figure}

\begin{figure}[tb] 
   \centering
   \includegraphics[width=0.95\columnwidth]{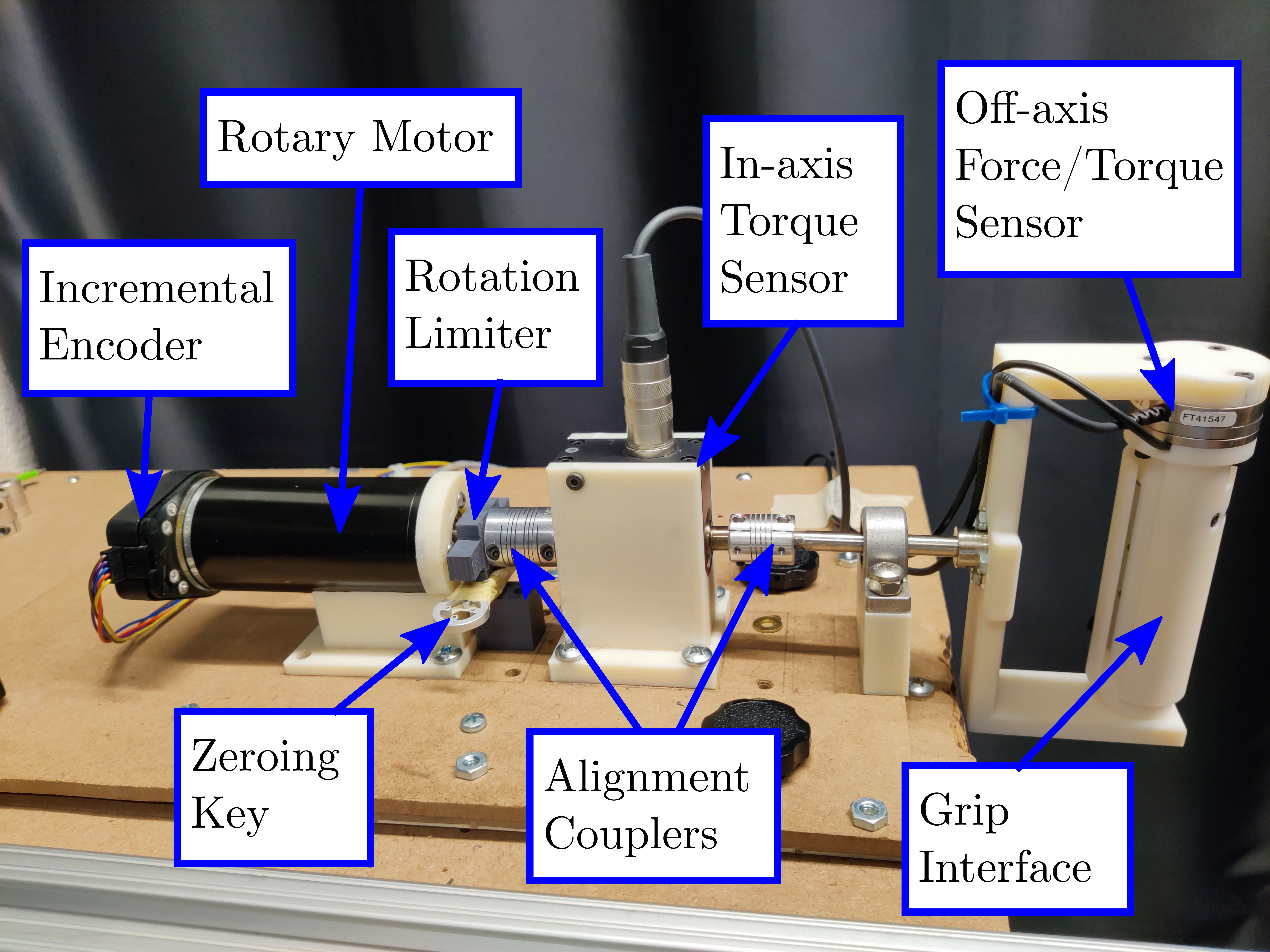}
   \caption{Side view of the robotic actuator of the WRIST testbed, with annotated callouts of key components.}
   \label{fig:wristparts}
\end{figure}

To collect psychometric data and present virtual tasks, we built and introduced the Wrist Impairment Device for Assessment and Treatment (WRIST) Testbed. A physical overview of the testbed assembly and electronic box is depicted in Figure~\ref{fig:wristsetup}, and a component breakdown of the testbed actuator is shown in Figure~\ref{fig:wristparts}. To render torques for the wrist, a Maxon RE50 motor is mounted to a rigid board and energized by a ESCON 70/10 amplifier. A US Digital E2 encoder is affixed to the motor shaft to measure rotational position. The motor is attached with an aluminum spiral-cut coupler to a Futek TRS600 torque sensor rated at 5\,N-m, used to monitor the difference in commanded torque and instructed input torque from the user. A custom PVC piece, known as the rotation limiter, is attached to the first coupler to limit the range of the device to a comfortable $\pm$60\,degrees. A zeroing key can be inserted underneath the rotation limiter to stabilize and zero the rotation of the device during initialization. A second coupler and 316 stainless steel rod connects the torque sensor to a grip interface for the user. The rod is supported with a mounted ball bearing, height-adjusted with a custom 3D-printed support underneath. The grip interface is attached to the end of the transmission rod with a flange-mount shaft collar. The rod is keyed so the set screw on the collar can align the interface assembly properly without risk of rotational shifting.

\begin{figure}[tb] 
   \centering
   \includegraphics[width=0.95\columnwidth]{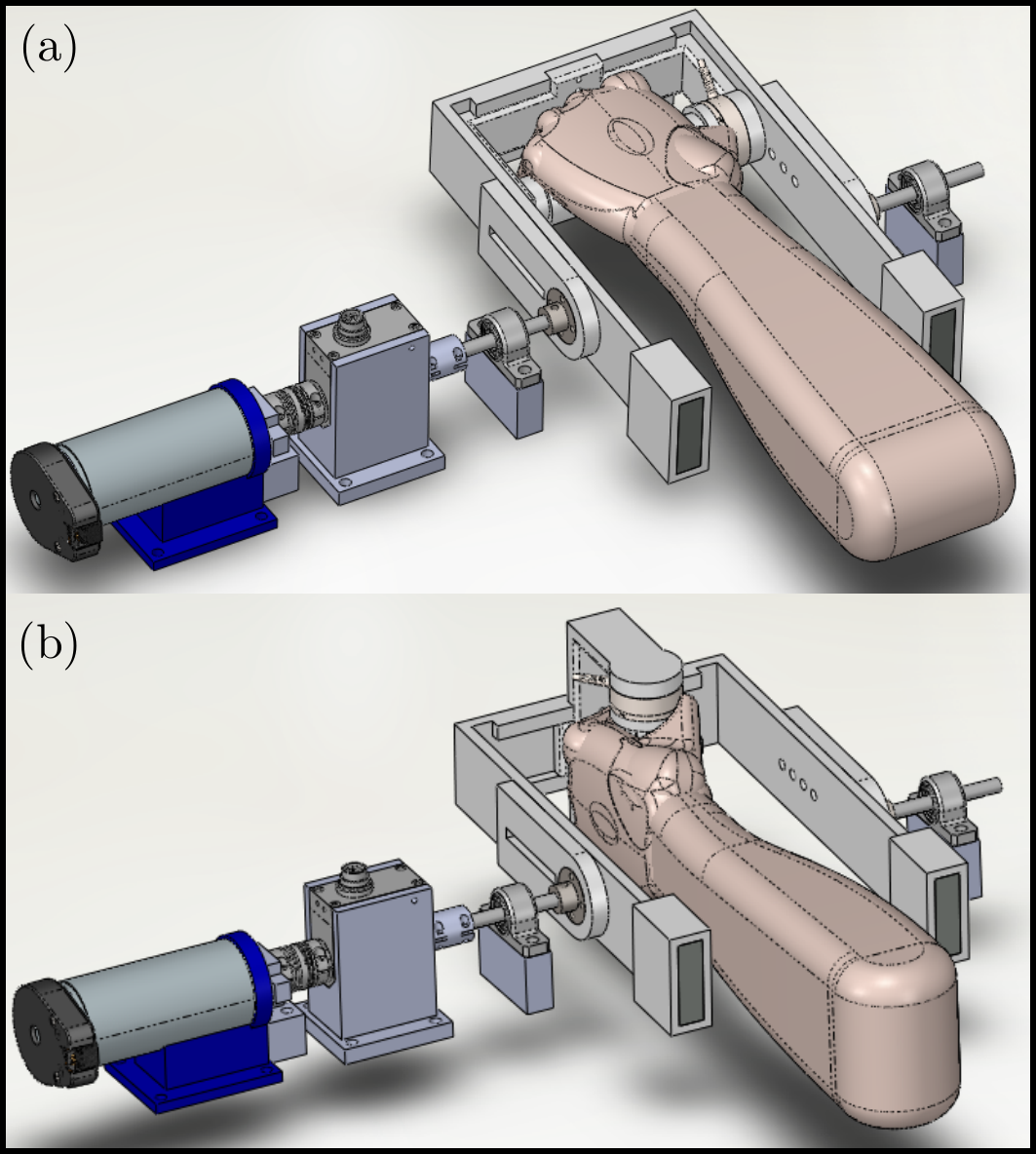}
   \caption{The WRIST Testbed utilizing different interface attachments to change and fix the direction of rotation for (a) flexion/extension and (b) abduction/adduction.}
   \label{fig:wristdirectioncasing}
\end{figure}

The grip interface consists of a tubular grip and a Y-handle. The tubular grip itself consists of two pieces of 3D-printed ceramic-plastic composite: a housing piece and a pressure piece. A LSP-10 beam load cell is fastened inside the cavity of the housing piece, and the pressure piece is attached separately to the load cell with no other mechanical connections. In this configuration, any squeezing or clenching of the hand can be captured as a force signal by the load cell. Positioning and fastening of the cylindrical force plate onto the load cell relative to the user’s hand was informed by prior literature on grasping distribution \cite{Seo2011EffectAbility}. One end of the hand grip assembly is attached to the Y-handle via an ATI Mini40 6-axis force-torque sensor, which is used to capture non-instructed off-axis forces and torque from the wrist. The Y-handle can be attached directly to the shaft collar for pronation/supination motion, but can also fit into an orthogonal casing to allow for either adduction/abduction motion or flexion/extension motion. 3D models of these configurations are shown in Figure~\ref{fig:wristdirectioncasing}.  All three anatomical axes of the wrist will be investigated due to the unique workspace and function provided by the joint in upper-limb manipulation.  For the purpose of this psychometric experiment, only the pronation/supination direction was concerned to test device and experiment feasibility in a straightforward manner.

\begin{figure}[tb] 
   \centering
   \includegraphics[width=0.95\columnwidth]{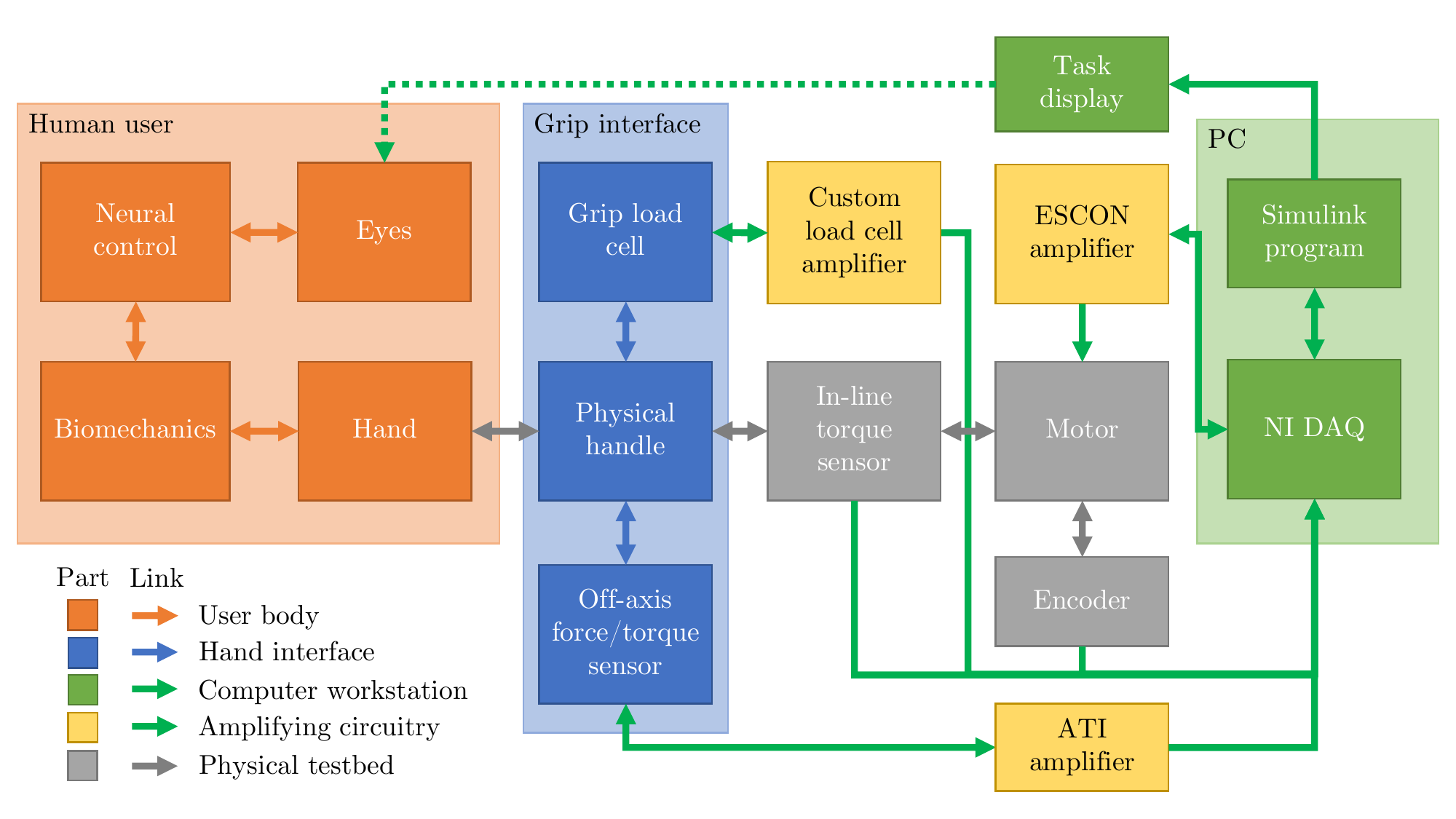}
   \caption{A graphical summary of all notable components of the human-device interaction system during a study using the WRIST testbed. Green arrows generally indicate electronic signal flows. Dashed arrows highlight visual connection that can be toggled.}
   \label{fig:wristsignal}
\end{figure}

\begin{figure}[tb] 
   \centering
   \includegraphics[width=0.95\columnwidth]{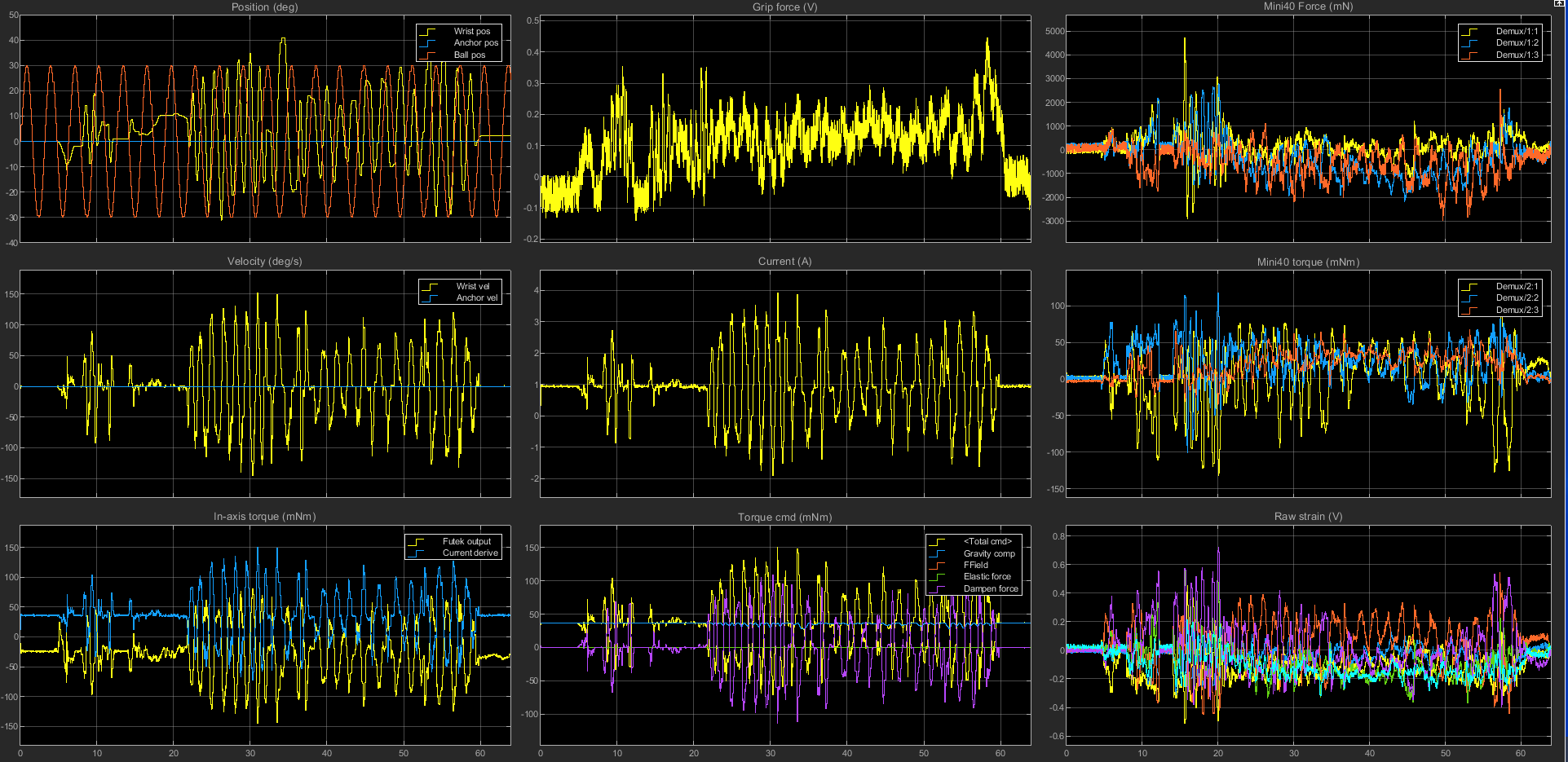}
   \caption{A scoped output of all signals processed by the Simulink software, which includes position, velocity, torque, grip force, and current.}
   \label{fig:wristscope}
\end{figure}

The sensors and motor are energized by and send signals to modules located inside an electronics housing. A flow chart of how these components interacted during user-study experimentation is illustrated in Figure~\ref{fig:wristsignal}. A custom amplification board with instrumental op-amp is used to increase the intensity of the grip force signals from the load cell. The ESCON 70/10 sends current commands to the Maxon motor, and sends back a readout signal of current consumed. A push-button emergency stop switch is connected to the enable port of the ESCON to stop unsafe or unwanted behavior by the investigator. Both the load cell amp board and ESCON are energized by a linear power supply, with the amp board needing a buck converter. An amplification box is connected to the Mini40 to condition the individual strain gauge signals. All signal lines are connected to a terminal board, which is directly connected to a National Instruments (NI) PCIe DAQ card inside a desktop PC. A MATLAB/Simulink program with Quanser's QUARC Real-Time Control Software is used to process signals to and from the DAQ card. Position signals from the motor encoder are differentiated by adjacent samples and filtered with 2nd-order Butterworth filters to derive stable velocity and acceleration signals. Force commands to the ESCON are influenced by virtual environmental forces (i.e.,~inertial, damping, or elastic), any gravity compensation for free-space motion, and custom control laws. A comprehensive display of received sensor signals and commanded actuation signals is illustrated in Figure~\ref{fig:wristscope}. 

\begin{figure}[tb] 
   \centering
   \includegraphics[width=0.95\columnwidth]{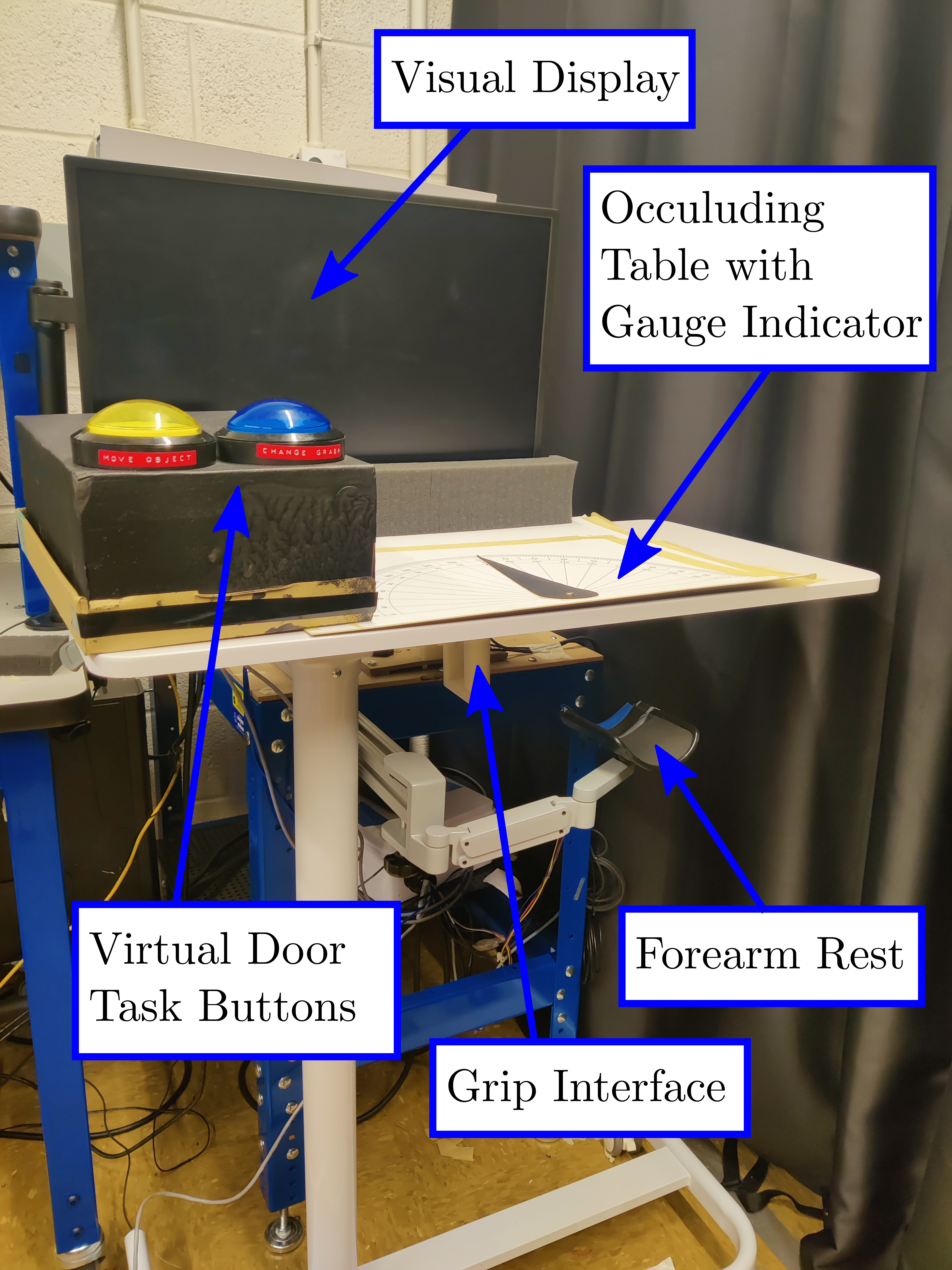}
   \caption{Front view of WRIST testbed with additional experimental equipment for user interaction and controlling visual confounds, along with annotated callouts.}
   \label{fig:wristdesk}
\end{figure}

For this experiment, additional equipment complemented the testbed to control user vision and hearing, shown in Figure~\ref{fig:wristdesk}. Visual renders of the virtual task environment were displayed to a high-definition monitor above the wrist device. Meanwhile, a height-adjustable desk could occlude the testbed from user sight, making it more difficult to see rotation and confound kinesthetic sensation with visual integration. Due to frequent switching between tasks with and without visual feedback, blindfolds were not implemented. Furthermore, noise-cancelling headphones were used to attenuate any distracting noise from the device mechanics or from the testing room.

\subsection{Participants}
We recruited N~=~11 participants (6 F) for our robotic assessment study. All participants were college-affliated adults (age: 29.5 $\pm$ 12.4). A tabular breakdown is shown in Table~\ref{tab:wristdemograph}. All participants were provided written informed consent according to a protocol approved by the Johns Hopkins School of Medicine Institutional Review Board (Study\# IRB00209185). Participants were compensated at a rate of \$10/hour.

\begin{table}[htbp]\setlength{\tabcolsep}{6mm}
  \centering
  \caption{Breakdown of age and gender across N~=~11 participants.}
  \begin{threeparttable}
    \begin{tabular}{lrl}
    \toprule
    PID\tnote{1}  & \multicolumn{1}{r}{Age} & \multicolumn{1}{l}{Gender} \\
    \midrule
    \multicolumn{1}{l}{1} & 25    & \multicolumn{1}{l}{M} \\
    \multicolumn{1}{l}{2} & 23    & \multicolumn{1}{l}{M} \\
    \multicolumn{1}{l}{3} & 19    & \multicolumn{1}{l}{F} \\
    \multicolumn{1}{l}{4} & 29    & \multicolumn{1}{l}{F} \\
    \multicolumn{1}{l}{5} & 25    & \multicolumn{1}{l}{M} \\
    \multicolumn{1}{l}{6} & 28    & \multicolumn{1}{l}{F} \\
    \multicolumn{1}{l}{7} & 44    & \multicolumn{1}{l}{M} \\
    \multicolumn{1}{l}{8} & 21    & \multicolumn{1}{l}{F} \\
    \multicolumn{1}{l}{9} & 21    & \multicolumn{1}{l}{F} \\
    \multicolumn{1}{l}{10} & 61    & \multicolumn{1}{l}{F} \\
    \multicolumn{1}{l}{11} & 28    & \multicolumn{1}{l}{M} \\
    \midrule
    Mean  & 29.45 &  \\
    STD\tnote{2}  & 12.43 &  \\
    Median & 25.00 &  \\
    \bottomrule
    \end{tabular}%
    \begin{tablenotes}
    \footnotesize
    \item[1]PID: Participant Identification number; \item[2]STD: Standard deviation
    \end{tablenotes}
     \end{threeparttable}
  \label{tab:wristdemograph}%
\end{table}%

\subsection{Study Design}
Upon confirmation of written informed consent, each participant was asked questionnaires regarding demographics and handedness (using the Edinburgh Handedness Inventory \cite{Oldfield1971TheInventory}). Functional assessments adapted from clinical settings were then conducted on the participant’s preferred side (for wrist pronation) to test and quantify proprioceptive and functional ability. The Nottingham Sensory Assessment with Erasmus MC modifications (emNSA) \cite{Stolk-Hornsveld2006TheDisorders} and the hand-wrist section of the Fugl-Meyer Assessment (FMA-HW) \cite{Fugl-Meyer1975ThePerformance.} exhibited good usability and robustness in prior literature \cite{Connell2012MeasuresReview}, so those were implemented in this study. Only the proprioceptive score protocol from emNSA was used. The last “pen-and-paper” clinical assessment was the Montreal Cognitive Assessment (MoCA) \cite{Nasreddine2005TheImpairment}, which was implemented to get an aggregate rating of different aspects of cognitive ability. 

\begin{figure}[tb] 
   \centering
   \includegraphics[width=0.95\columnwidth]{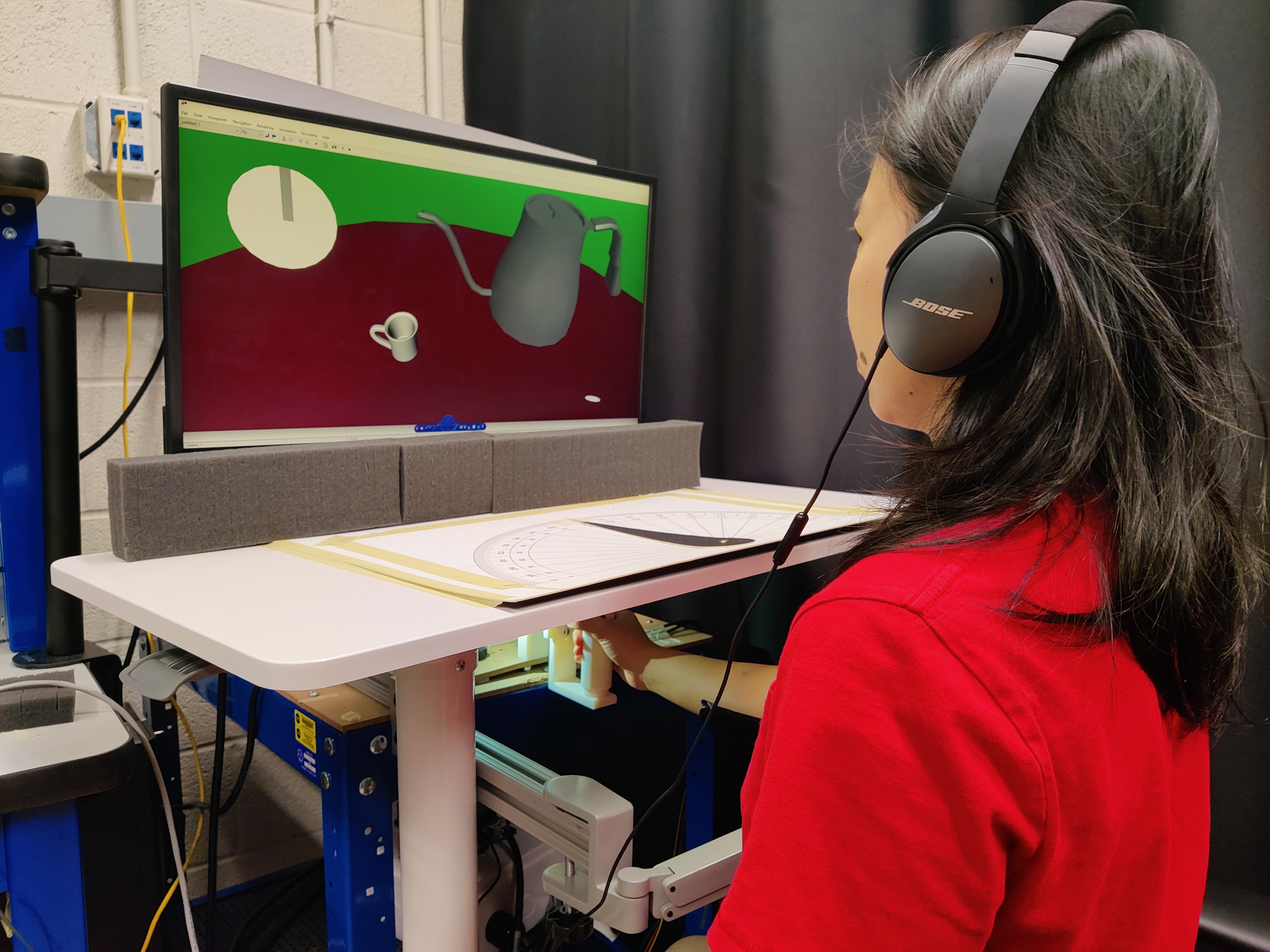 }
   \caption{Typical human-robot interfacing during an virtual assessment task with display monitor output. The user’s forearm was supported to mitigate unnecessary fatigue. Note the table, foam, and monitor blocked user sightlines to the rotating equipment to control the influence of vision, and the headphones blocked out distracting noise.}
   \label{fig:wristinterface}
\end{figure}

The participant was then adjusted in a chair and grips the interface of the device, as shown in Figure~\ref{fig:wristinterface}. The user pronated as far as they can comfortably, or until they hit the rotation limiter, and then supinated as far as they could comfortably, or until they hit the limiter. The difference between these two angular limits defined the user's comfortable range of motion (cROM). This limited variant of range of motion was used instead of larger anatomical or trainable versions to ensure the safety of the user in a preliminary robotic implementation.

\begin{figure}[tb] 
   \centering
   \includegraphics[width=0.95\columnwidth]{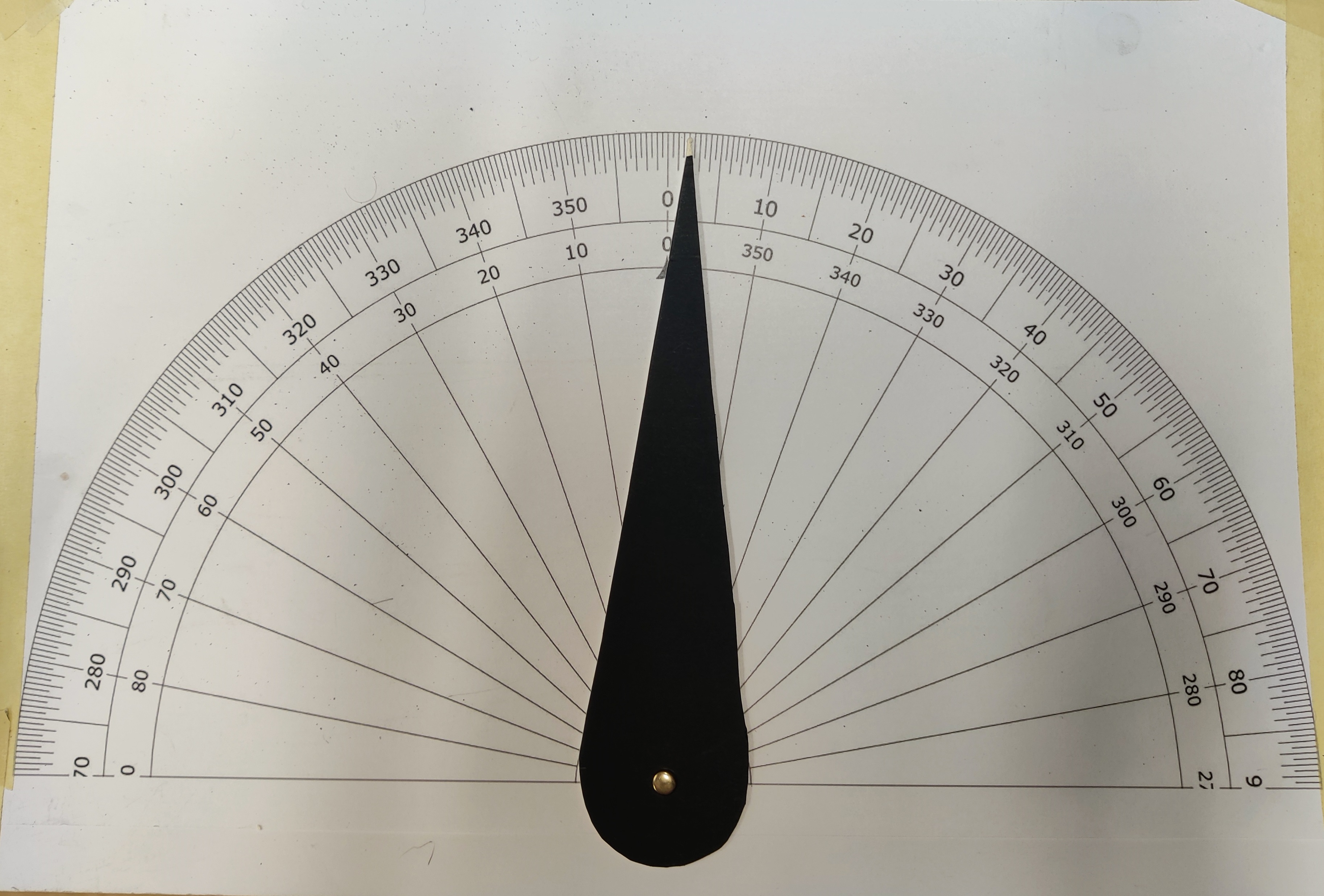}
   \caption{The physical angular indicator on the user table utilized in the passive positional gauge matching task. The pointer can freely pivot from user input to a particular angle (in degrees) to match the angle of the user’s hand controlled by the robot.}
   \label{fig:wristgauge}
\end{figure}

A robotic version of a passive gauge matching paradigm with good test-retest reliability \cite{Rinderknecht2016ReliableParadigm}, in turn an adaptation of the Wrist Position Sense Test \cite{Carey1996ImpairedUse}, was utilized to measure position sense in contralateral pointing. This test involved representing a wrist rotation by adjusting a physical indicator with the other hand in an environment adjacent to the user (i.e.,~extrapersonal space). Specifically, a sequence of robotic rotations to different angles in the user’s cROM was performed, and users had to indicate the angle their hand rotated to. Users indicated by pivoting a gauge pointer on a large protractor with degree resolution seen in Figure~\ref{fig:wristgauge}, and with verbal confirmation if possible. The difference between hand position and user-reported position was recorded for 20 unique angle presentations across the cROM. No visual display from the monitor was used for this test.

\begin{figure}[tb] 
   \centering
   \includegraphics[width=0.95\columnwidth5]{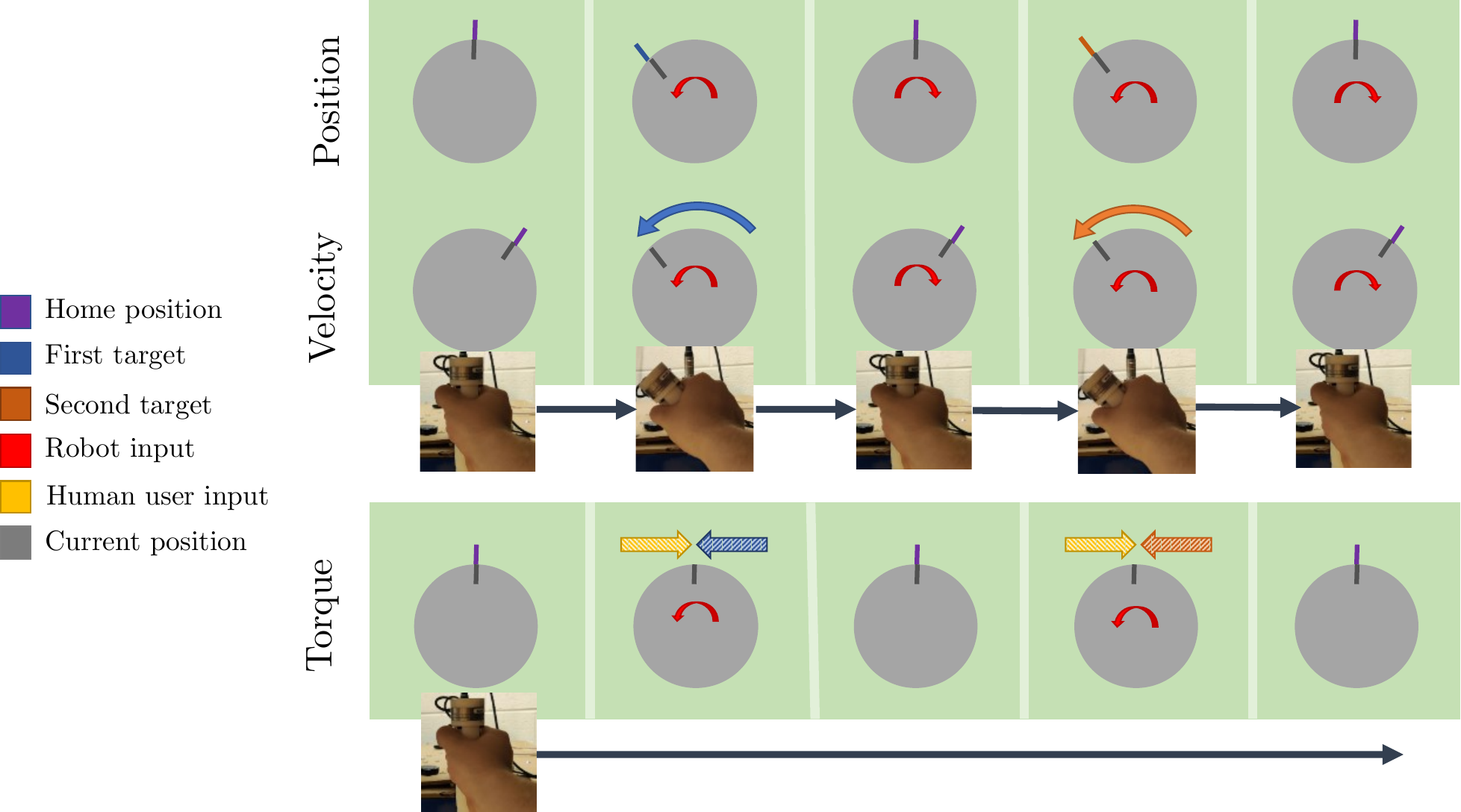}
   \caption{Presentation progression (left-to-right) of rotational discrimination task for a 2-interval forced choice of different wrist precepts. Light gray circles represent the wrist interface, interior ticks represents the interface position, and exterior ticks represent a presented stimulus. Curved arrows near the interface center indicate proactive motor command torques, curved arrows around the perimeter indicate angular velocities, and tangent straight arrows indicate the force balance between stimulus torque and reactive torque from the user.}
   \label{fig:wristdisrim}
\end{figure}

\begin{figure}[tb] 
   \centering
   \includegraphics[width=0.95\columnwidth]{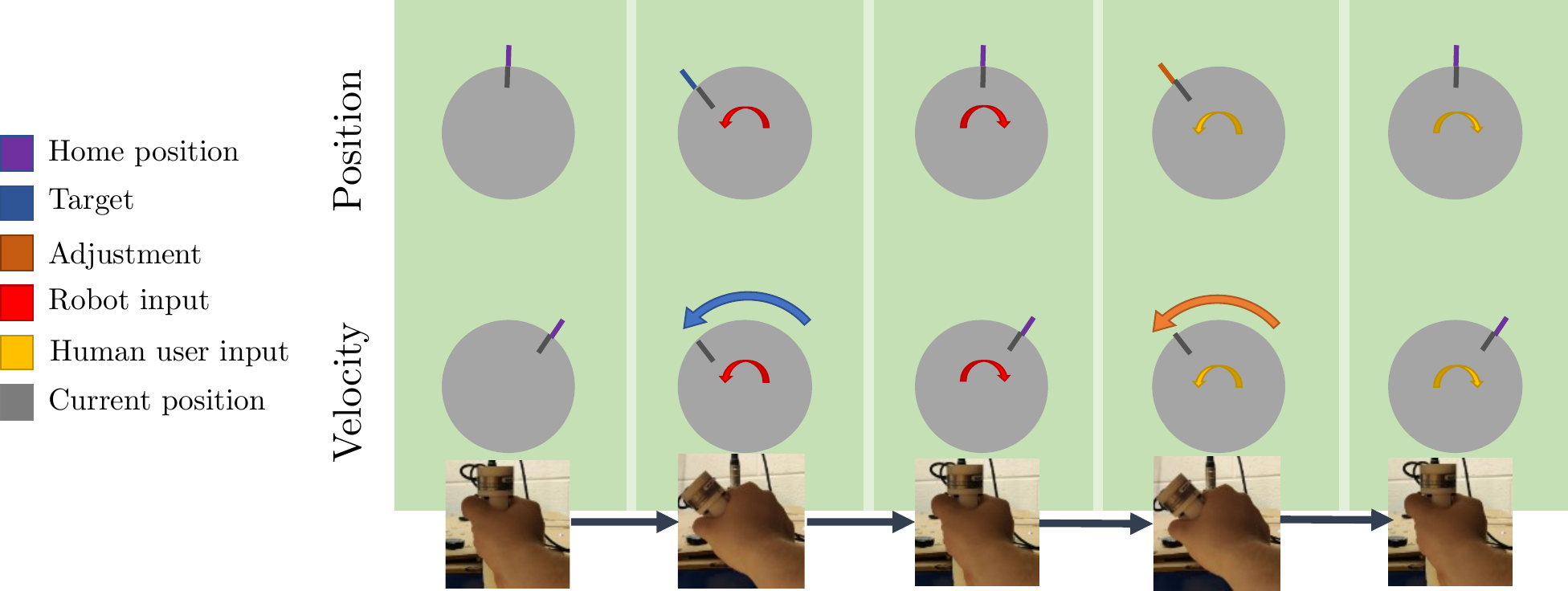}
   \caption{Presentation progression (left-to-right) of rotational task where users had to match a presented stimulus afterwards for different wrist precepts. Light gray circles represent the wrist interface, interior ticks represents the interface position, and exterior ticks represent a presented stimulus. Curved arrows near the interface center indicate proactive command torques from either the robot or user and curved arrows around the perimeter indicate angular velocities.}
   \label{fig:wristadjust}
\end{figure}

The robotic tasks (stimuli discrimination, stimuli adjustment/reproduction, and virtual ADL) were represented as seven blocks that were counterbalanced across participants. For both discrimination and reproduction task blocks (referred generally as psychometric tests), there was a block representing position and velocity each for both test types, and torque in addition for discrimination. For visual reference, discrimination tasks are illustrated in Figure~\ref{fig:wristdisrim}, and adjustment tasks are illustrated in Figure~\ref{fig:wristadjust}.

\begin{enumerate}
\item Position discrimination (passive-passive):
For each trial, a reference position was set as 30\% of the user’s cROM in pronation. Using a 3-down-1-up (3D1U) transformed weighted adaptive staircase, the comparison position was generated based on the reference position. The presentation order between the two positions was randomized. The device was driven to a home neutral position near zero. The participant was instructed not to input torque. The device rotated the wrist in pronation to the first position, held by stiff virtual walls. The position was presented for 2\,seconds, then the wrist was rotated back to home position. After 1.75\,seconds, the wrist was rotated to the second position, held for 2\,seconds, and returned to home. The participant reported to the investigator if the second position differed from the first. The response of the participant was used to update the staircase algorithm to inform the next comparison position. The section ended when the maximum trial count of 50 was hit or the just-noticable difference (JND) reversal condition of 8 reversals was satisfied.
\item Velocity discrimination (passive-passive):
For this task, reference velocity was set at 60\,degrees per second. Using a 3D1U adaptive staircase, the comparison velocity was generated based on the reference velocity. The presentation order between the two speeds was randomized. The device was driven to a home position at the user’s neutral angle. The participant was instructed not to input torque. The device rotated in pronation at the first velocity, guided by a gap between a moving set of virtual walls. After the interface reached the target position of 30\% cROM, it was held for 2\,seconds then rotated the wrist back to home position. After 1.75\,seconds, the wrist was rotated at the second velocity to the target position, held, and returned home. The participant reported to the investigator if the second velocity differed from the first. The response of the participant was used to update the staircase algorithm to inform the next comparison velocity. The section ended when the maximum trial count of 50 was hit or the JND reversal condition of 8 reversals was satisfied.
\item Torque discrimination (reactive-reactive):
For this task, the reference torque was set at 500\,mNm to protect the user and device. Using a 3D1U adaptive staircase, the comparison torque was generated based on the reference torque. The presentation order between the two torques was randomized. The device was driven to a zero-rotation (or home) position.  The participant was instructed not to input torque until prompted verbally and the motor began the torque ramp-up. The participant attempted to counter-act the torque and maintain home position for the duration of the robotic torque. The first torque then stopped. After 1.75\,seconds, the second torque was presented for the user to act against, and then stopped. The participant reported to the investigator if the second torque differed from the first. The response of the participant was used to update the staircase algorithm to inform the next comparison torque. If the user failed to maintain position within the tolerance region during the trial, that trial’s data was ignored, and the adaptive staircase was not adjusted.
\item Position adjustment (passive-proactive):
For each trial, a reference position was selected randomly in the range of pronation motion. The device was driven to a zero-rotation (or home) position. The participant was instructed not to input torque. The device rotated the wrist to the first position, held by stiff virtual walls. The position was presented for 2\,seconds, then the wrist was rotated back to home position. The return to home position is important, given that position coding in healthy individuals tend to reproduce relative amplitudes rather than absolute positions \cite{Marini2017CodificationSense}. After 1.75\,seconds, the robot stopped rendering walls to allow for free exploration, and the investigator then instructed the participant to reach the reference position and return to the neutral position. After user replication, the robot re-rendered the walls to steady the user for the next angle presentation. The section ended when the maximum trial count of 21 angle presentations was hit.
\item Velocity adjustment (passive-proactive):
For each trial, a reference velocity was selected randomly in a range of acceptable speeds with a given direction. The device started at home position at a neutral angle. The participant was instructed not to input torque. The device rotated at the first velocity, guided by a gap between a moving set of virtual walls. After the interface reached an end position of 30\% cROM pronation, then the wrist was rotated back to home position. The robot stopped rendering walls to allow for free exploration, and the investigator then instructed the participant to replicate the reference velocity consistently to the end position and back. After user replication, the robot re-rendered the walls to steady the user for the next velocity presentation. The section ended when the maximum trial count of 21 velocity presentations was hit.
\item Kettle pouring (ADL):
In this virtual ADL task, participants had to pour tea from a kettle into a cup on a table by tilting it at the correct angle, which is shown in Figure~\ref{fig:wristkettleflow}. The purpose of the virtual ADL task was to understand completion speed of a timed holistic task found in everyday living. With the rich quantitative data of real-time motion data in this task, kinematic analysis can be also conducted in later studies to better understand human motion control.  The rotation of the interface determined the tilt of the kettle, which determined the flow rate of liquid into the cup. The user had to fill the cup to a specified level in a certain amount of time, which will result in a successful trial. If the cup was underfilled after the allotted time or was overfilled, then the task will result in a failed trial.

\begin{figure}[tb] 
   \centering
   \includegraphics[width=0.95\columnwidth]{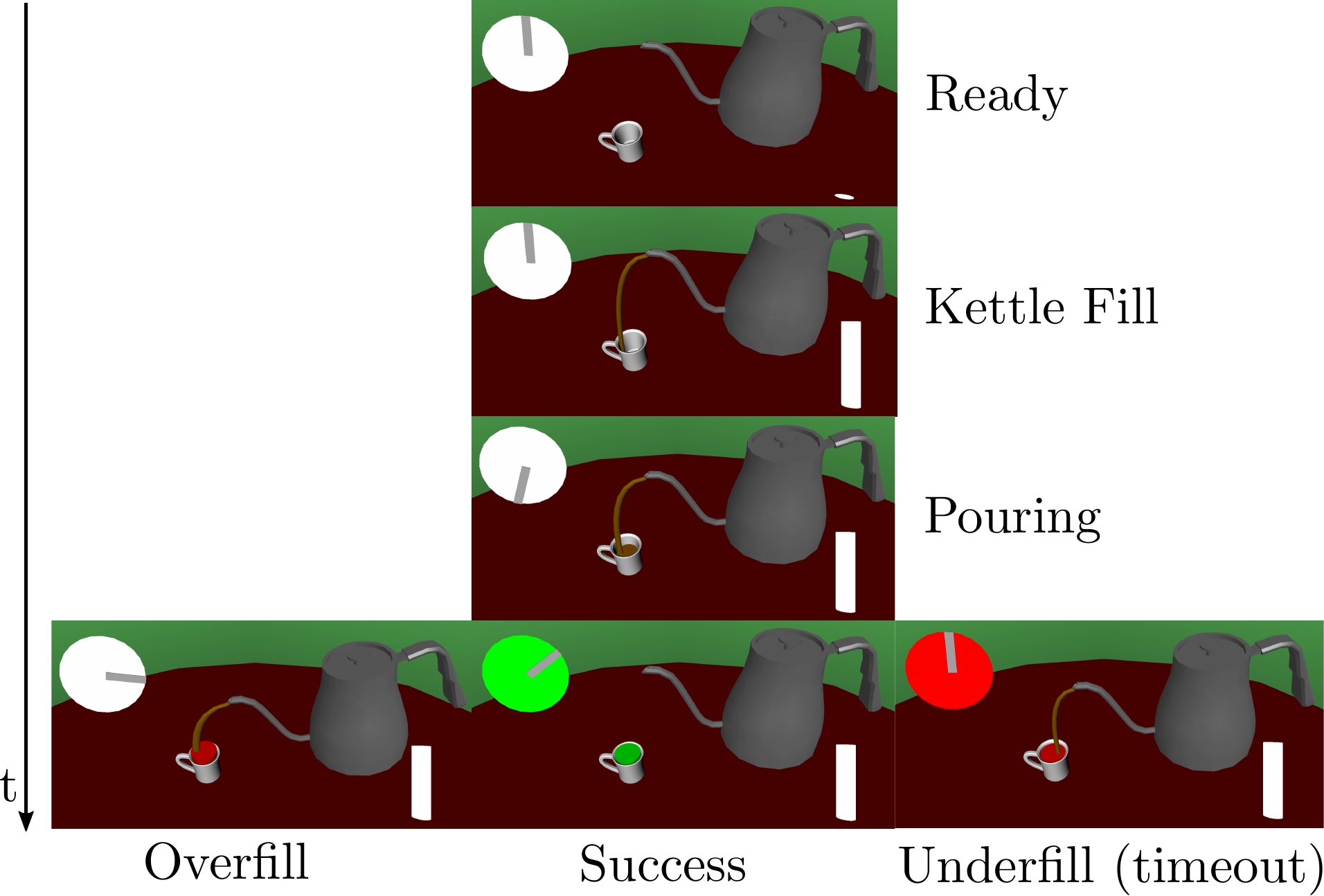}
   \caption{The progression of the kettle task over time is shown from top to bottom. The last state at bottom represents different success and failure conditions depending on how well the user controlled the fill line of the liquid in the cup within the allotted time limit.}
   \label{fig:wristkettleflow}
\end{figure}

\item Door opening (ADL):
In this virtual ADL task, participants had to check if a door was locked with a hand avatar represented by a blue glove, and subsequently open it via key insertion and turning. The task flow is detailed in Figure~\ref{fig:wristdoorflow}. The purpose of this virtual ADL task was to investigate how quick people can perform a sequence of actions to accomplish a goal in everyday living without time constraints. Hand movement and object pushing/pulling were abstracted to push buttons separate from the grip interface, shown in Figure~\ref{fig:wristbuttons}, to explore increased task difficulty and to explore the feasibility of contralateral user interfaces. Users first had to grab onto and rotate a doorknob to check if the door was locked. Then, users shifted their hand near a key to grab it and align it with the doorknob’s keyhole. After inserting the key, the user had to rotate the key to unlock the deadbolt. Finally, users had to grab and twist the doorknob so that the door could be opened.

\begin{figure}[tb] 
   \centering
   \includegraphics[width=0.9\columnwidth]{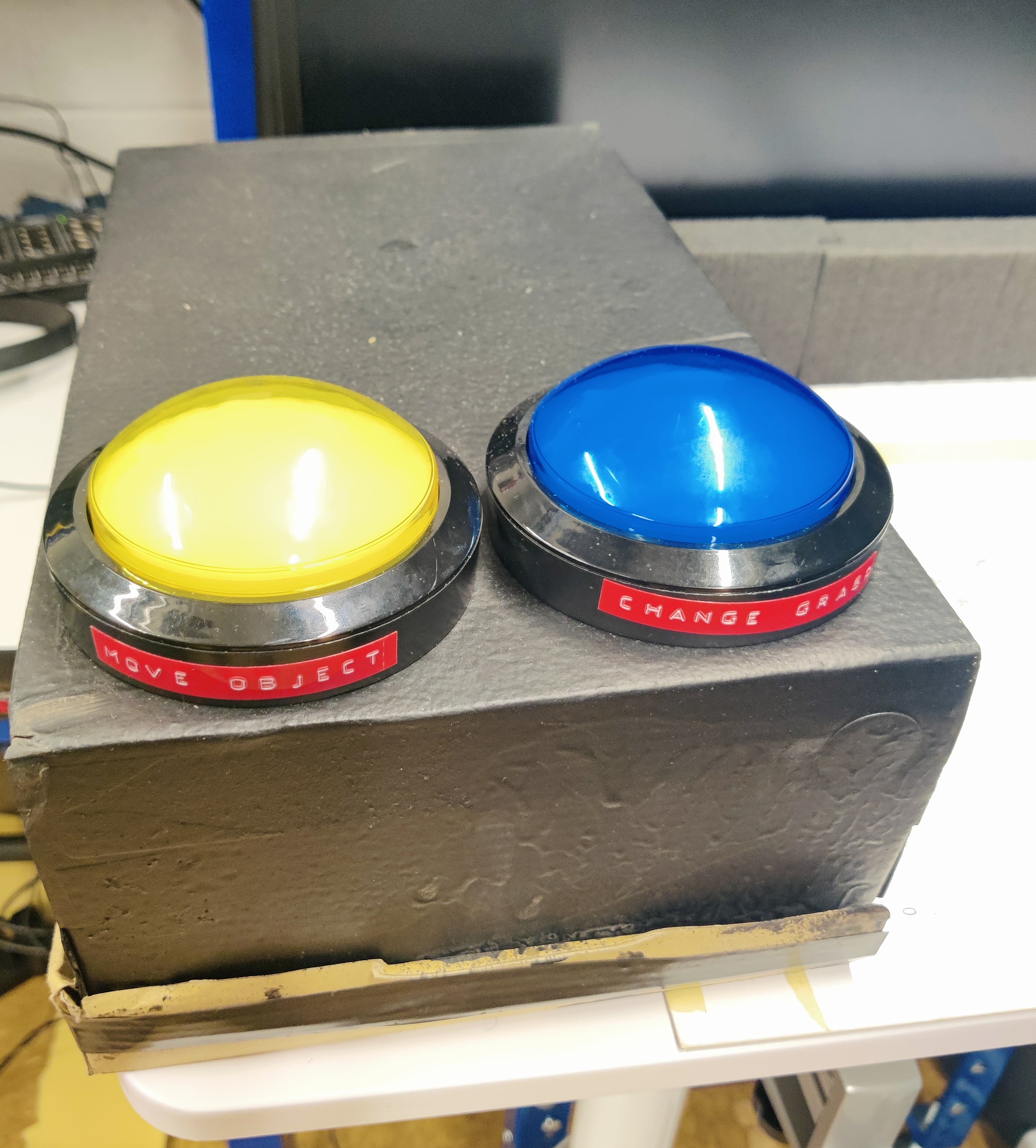}
   \caption{The contralateral button interface used by the user to specify motion of objects towards and away from the virtual door. The blue button just moved the posture of the blue gloved hand towards and away from the door and releasing any grasped object, and the yellow button caused hand movement to push or pull any grasped object with the hand avatar.}
   \label{fig:wristbuttons}
\end{figure}

\begin{figure}[tb] 
   \centering
   \includegraphics[width=0.95\columnwidth]{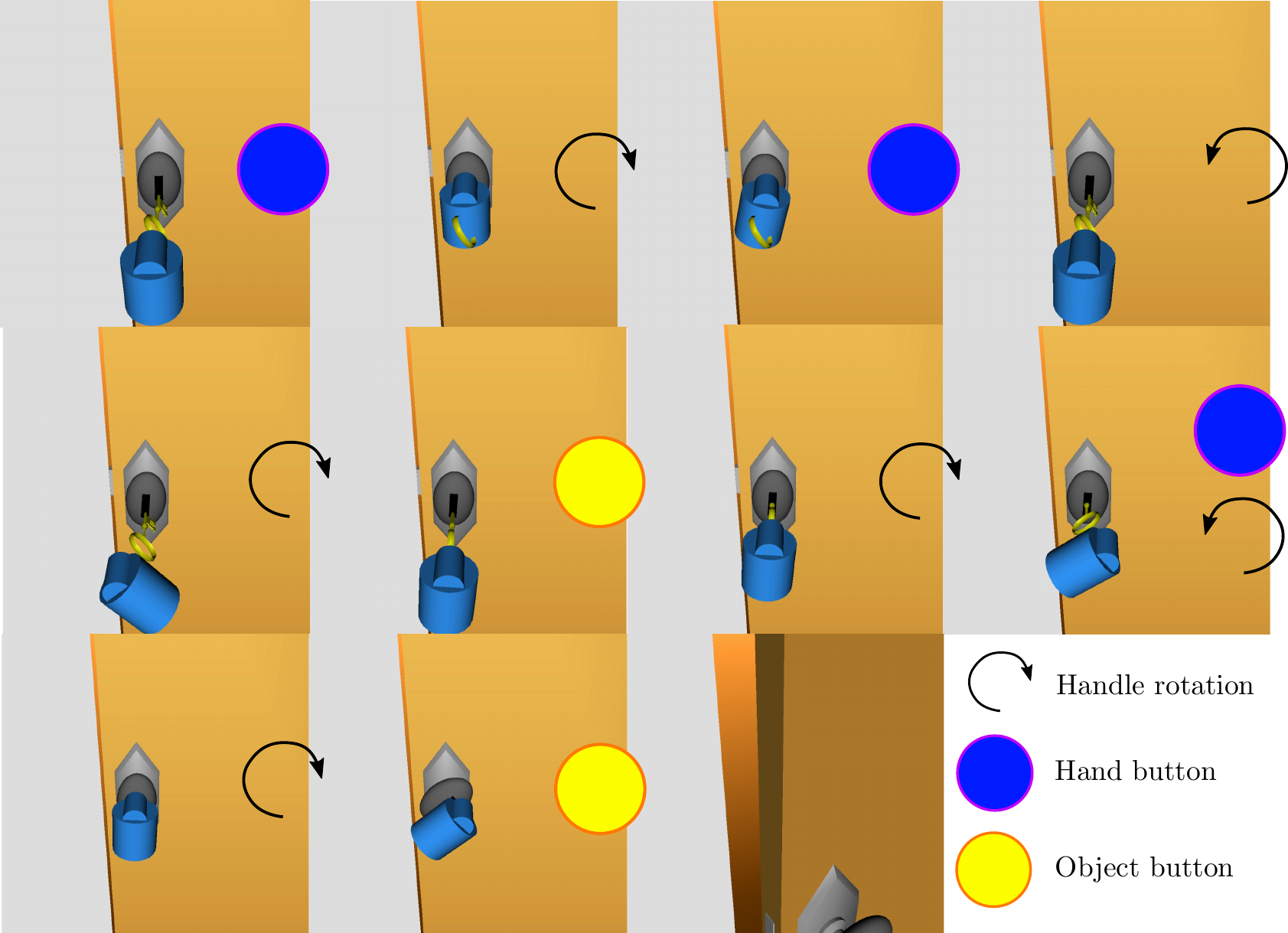}
   \caption{The step-by-step progression of door exploration during the door opening task. Progression is left-to-right, top-to-bottom. Icons are placed between depicted states to indicate user actions to progress the task. Colored circles represent user input via the corresponding physical button, and the circular arrows indicate user input via the rotation of the grip interface.}
   \label{fig:wristdoorflow}
\end{figure}

\end{enumerate}

\begin{figure*}[tb] 
   \centering
   \includegraphics[scale=0.5]{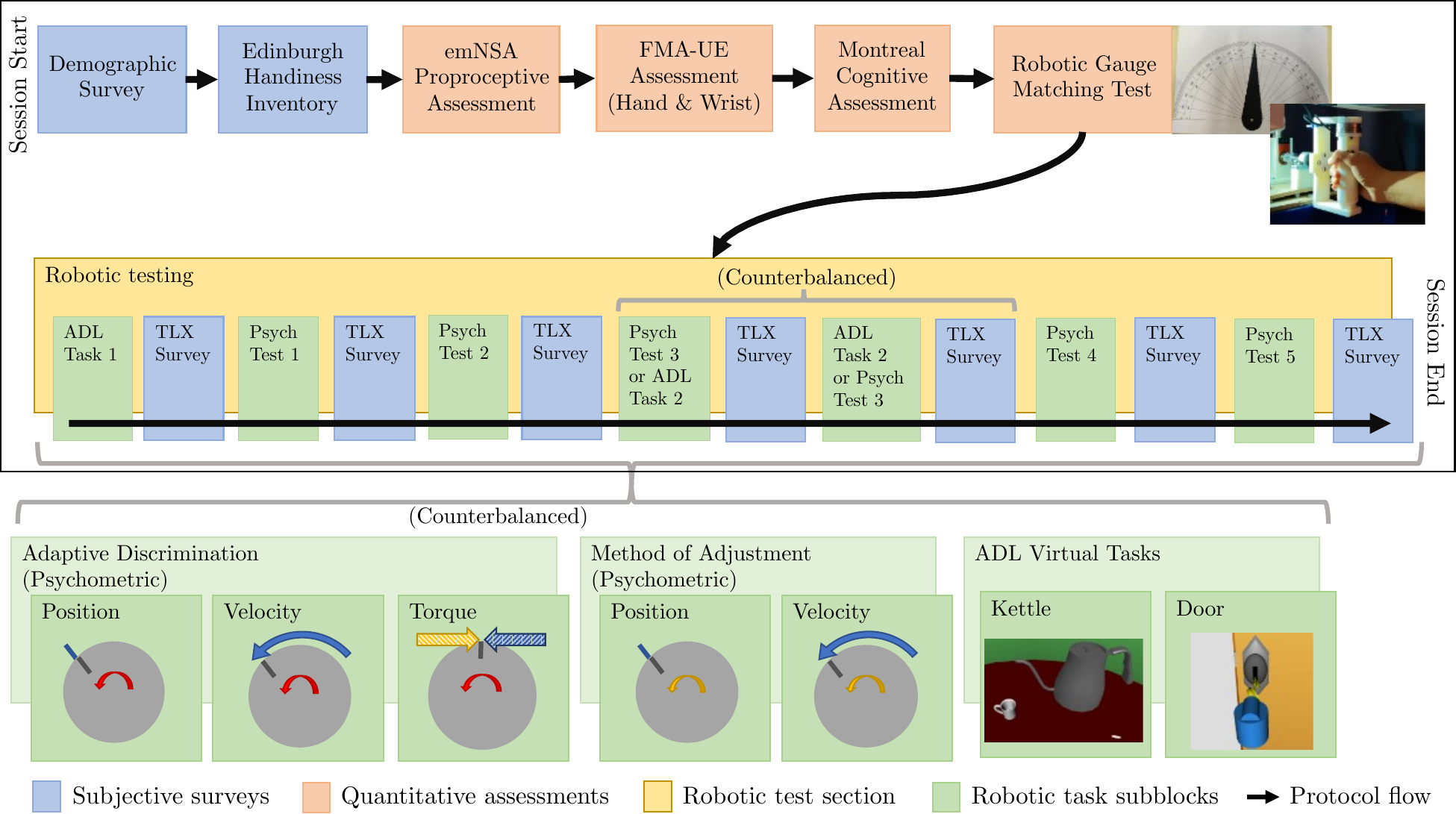}
   \caption{The protocol progression for a typical session of psychometric and ADL assessment on the WRIST testbed. Quantitative assessments and questionnaires were used to rate components of functional and cognitive ability.}
   \label{fig:wristprotocol}
\end{figure*}

The overall structure of the protocol incorporating these tasks is illustrated in Figure~\ref{fig:wristprotocol}. To account for potential ordering effects of task presentation, a pseudo-random counterbalancing was performed across all participants. To maintain user attention and interest, the ADL tasks were specifically placed after gauge matching and in the middle of the robotic testing section. Due to an even number of tests, the 2nd ADL and 3rd psychometric tests were also pseudo-random counterbalanced across participants, so the 2nd ADL approximately represents the middle of robotic testing across the sample.

\subsection{User-reported Surveys}
Participants were asked to self-report age, gender, and handedness for a demographic survey before the clinical assessments. After each robotic task, participants responded to six questions from the NASA-TLX (Task Load Index) questionnaire \cite{Hart1988DevelopmentResearch} regarding perceived effort and comfort. Specifically, they were asked about mental demand, physical demand, temporal demand, performance, effort, and frustration.

\subsection{Metrics and Statistical Analysis}
Each task on the Edinburgh Handedness Inventory was scored based on how exclusively the participant's left or right hand was used to perform activities of daily living, as reported by the participant. The resulting handedness scores were normalized into a (right) laterality index scaled from -100 to 100. The proprioception subscore of emNSA was scored on a scale of 0 to 8 for each participant. The hand-wrist subscore of FMA-UE was scored on scale from 0 to 30. Scores from the MoCA were scaled from 0 to 30. Each NASA-TLX question was answered on a scale from 0 to 100.

Discrimination tasks reported a JND as an absolute value and a percentage of reference stimuli (a Weber fraction as Kp, Kv, or Kt for different affects). For each ispilateral discrimination task, the JND was taken as the median of the comparison stimuli for the last four reversals of the adaptive staircase. Positional JND is abbreviated as JNDp, velocity JND is abbreviated as JNDv, and torque JND is abbreviated as JNDt. 

Ipsilateral adjustment tasks reported errors between reference stimuli and user matching input from the same hand. Meanwhile, contralateral gauge matching tasks reported errors between the angle presented by the robot to the passive wrist of interest and the user-reported angle indicated on the gauge by the other (active) hand. Gauge matching and positional reproduction errors are reported as an average across trials in degrees, determined by the difference in steady-state position between reference angle and user-produced angle. Gauge matching error is abbreviated as MEg, and positional matching error is abbreviated as MEp.

Velocity reproduction errors are reported as an average across trials in degrees/second, determined by the difference in average rate of angle change between rising ramps of the reference rotation and the user-produced rotation. The ramp interval of interest was shortened to a subset midrange, as done in previous literature \cite{Nagai2016ConsciousSubmodality}, to cut off smoothstep transitions or user overshoot that might impact feature extraction. Velocity matching error is abbreviated as MEv.

Measures from the kettle pouring ADL virtual task included trial status (i.e.,~success, overfilled, underfilled) if different between participants, and completion time (Tk) for successful trials. The primary measure from the door status was completion time (Td).

All statistical analysis was performed using the Statistics and Machine Learning Toolbox in MATLAB 2021a (Mathworks; Natick, MA). Linear correlation trends between assessment scores were computed across the sample using Spearman’s non-parametric rank correlation for non-normal samples, although Pearson correlation may be used if normality is confirmed. As mentioned before, a rank correlation coefficient computation was selected to identify associations between scores, so that high (or low) performance in one score may predict high (or low) performance in another score.

As previously stated, scores in position or velocity senses may vary between tasks with different contexts. To check for these potential differences between measures of similar psychometric tests for a given sensory modality, a series of variance analysis tests was used. For the JND measure from velocity discrimination (JNDv) and the ME measure from velocity reproduction (MEv), a non-parametric Wilcoxon signed rank test was used, but a t-test for matched pairs can be used if normality is confirmed in the measure samples. For the ME measures from contralateral gauge matching (MEg) and ipsilateral position reproduction (MEp) and the JND measure for ipsilateral position discrimination (JNDp), a non-parametric Friedman test with Bonferroni correction was used, but a repeated-measures ANOVA can be used if normality of all samples exist.

\section{Results}
\subsection{Measures of Psychometric Acuity Generally Do Not Correlate Strongly in Rank with Each Other}
Except for the kettle and door ADL time measures, all measure samples were not normally distributed; therefore, non-parametric tests were conducted. Scoring breakdown between participants for all clinical tests are listed in Table~\ref{tab:wristclinic}.  All participants had perfect emNSA and FMA-HW scores, so they were excluded from correlation analysis.  Rotational workspaces of each participant captured by the gauge test are characterized in Table~\ref{tab:wristROM}. Robotic test scores are listed in Table~\ref{tab:wristscores}. All participants completed the ADL tasks with sufficient success, so binary success status was not reported.

\begin{table}[tb]
  \centering
  \caption{Scores from clinical assessments, including Edinburgh Handedness Inventory, Nottingham Sensory Assement, Fugl-Meyer Assessment, and Montreal Cognitive Assessment.}
    \begin{tabular}{lrrrr}
    \toprule
    PID   & \multicolumn{1}{r}{Handedness} & \multicolumn{1}{r}{emNSA} & \multicolumn{1}{r}{FMA-HW} & \multicolumn{1}{r}{MoCA} \\
    \midrule
    \multicolumn{1}{l}{1} & 26.32     & 8     & 30    & 26 \\
    \multicolumn{1}{l}{2} & 100.00    & 8     & 30    & 28 \\
    \multicolumn{1}{l}{3} & 100.00    & 8     & 30    & 30 \\
    \multicolumn{1}{l}{4} & 100.00    & 8     & 30    & 30 \\
    \multicolumn{1}{l}{5} & 20.00     & 8     & 30    & 29 \\
    \multicolumn{1}{l}{6} & 76.47    & 8     & 30    & 26 \\
    \multicolumn{1}{l}{7} & 100.00    & 8     & 30    & 25 \\
    \multicolumn{1}{l}{8} & 100.00    & 8     & 30    & 29 \\
    \multicolumn{1}{l}{9} & 100.00    & 8     & 30    & 29 \\
    \multicolumn{1}{l}{10} & -40.00    & 8     & 30    & 28 \\
    \multicolumn{1}{l}{11} & -20.00    & 8     & 30    & 28 \\
    \midrule
    Mean  & 60.25 & 8.00  & 30.00 & 28.00 \\
    STD   & 53.86  & 0.00  & 0.00  & 1.67 \\
    Median & 100.00 & 8.00  & 30.00 & 28.00 \\
    \bottomrule
    \end{tabular}%
  \label{tab:wristclinic}%
\end{table}%

\begin{table}[tb]
  \centering
  \caption{The ranges of motion in pronation/supination and the neutral home angle relative to robot zero position reported as comfortable by participants.}
    \begin{tabular}{lrr}
    \toprule
    PID   & \multicolumn{1}{r}{Neutral (deg)} & \multicolumn{1}{r}{cROM (deg)} \\
    \midrule
    \multicolumn{1}{l}{1} & 4.40  & 110.90 \\
    \multicolumn{1}{l}{2} & 1.10  & 119.50 \\
    \multicolumn{1}{l}{3} & -2.00 & 118.50 \\
    \multicolumn{1}{l}{4} & 1.50  & 118.50 \\
    \multicolumn{1}{l}{5} & -2.40 & 118.90 \\
    \multicolumn{1}{l}{6} & 0.90  & 118.80 \\
    \multicolumn{1}{l}{7} & 14.00 & 110.80 \\
    \multicolumn{1}{l}{8} & 0.90  & 118.60 \\
    \multicolumn{1}{l}{9} & 3.10  & 118.40 \\
    \multicolumn{1}{l}{10} & -2.60 & 118.70 \\
    \multicolumn{1}{l}{11} & -1.70 & 108.00 \\
    \midrule
    Mean  & 1.56  & 116.33 \\
    STD   & 4.72  & 4.20 \\
    Median & 0.90  & 118.50 \\
    \bottomrule
    \end{tabular}%
  \label{tab:wristROM}%
\end{table}%

\begin{table*}[tb]
  \centering
  \caption{Psychometric and time duration test measures from the robotic discrimination, adjustment, and virtual ADL assessments.}
    \begin{tabular}{lrrrrrrrrrrr}
    \toprule
    \multicolumn{1}{L{3.5em}}{PID} & \multicolumn{1}{R{3em}}{MEg (deg)} & \multicolumn{1}{R{3em}}{JNDp (deg)} & \multicolumn{1}{R{3.2em}}{Kp (\%)} & \multicolumn{1}{R{3.2em}}{JNDv (deg/s)} & \multicolumn{1}{R{3.2em}}{Kv (\%)} & \multicolumn{1}{R{3.7em}}{JNDt (mNm)} & \multicolumn{1}{R{3.2em}}{Kt (\%)} & \multicolumn{1}{R{2.5em}}{MEp (deg)} & \multicolumn{1}{R{3.3em}}{MEv (deg/s)} & \multicolumn{1}{R{3em}}{Tk (s)} & \multicolumn{1}{R{3em}}{Td (s)} \\
    \midrule
    \multicolumn{1}{l}{1} & 11.13 & 4.74  & 14.25 & 9.94  & 16.57 & 59.65 & 11.93 & 4.69  & 21.01 & 13.96 & 59.31 \\
    \multicolumn{1}{l}{2} & 5.33  & 1.52  & 4.25  & 6.46  & 10.77 & 137.10 & 27.42 & 4.69  & 18.52 & 6.35  & 22.02 \\
    \multicolumn{1}{l}{3} & 6.07  & 4.41  & 12.42 & 5.95  & 9.92  & 46.25 & 9.25  & 7.03  & 13.28 & 5.71  & 21.01 \\
    \multicolumn{1}{l}{4} & 6.35  & 1.51  & 4.25  & 7.45  & 12.42 & 82.85 & 16.57 & 5.22  & 16.44 & 6.60  & 23.59 \\
    \multicolumn{1}{l}{5} & 7.30  & 5.80  & 16.25 & 14.26 & 23.76 & 144.45 & 28.89 & 7.81  & 22.59 & 10.38 & 34.57 \\
    \multicolumn{1}{l}{6} & 10.53 & 2.17  & 6.08  & 10.56 & 17.60 & 138.00 & 27.60 & 5.14  & 17.77 & 6.70  & 46.16 \\
    \multicolumn{1}{l}{7} & 10.77 & 8.34  & 25.09 & 9.65  & 16.08 & 112.75 & 22.55 & 6.46  & 13.85 & 6.16  & 92.32 \\
    \multicolumn{1}{l}{8} & 13.75 & 2.23  & 6.26  & 7.16  & 11.93 & 91.10 & 18.22 & 3.69  & 9.49  & 11.93 & 21.22 \\
    \multicolumn{1}{l}{9} & 12.87 & 1.51  & 4.25  & 12.35 & 20.59 & 59.65 & 11.93 & 6.75  & 13.54 & 6.07  & 21.57 \\
    \multicolumn{1}{l}{10} & 5.46  & 4.12  & 11.57 & 4.96  & 8.27  & 149.60 & 29.92 & 3.34  & 9.47  & 6.11  & 56.38 \\
    \multicolumn{1}{l}{11} & 8.66  & 5.17  & 15.95 & 2.77  & 4.61  & 77.95 & 15.59 & 3.36  & 18.06 & 13.86 & 62.66 \\
    \midrule
    Mean  & 8.93  & 3.77  & 10.96 & 8.32  & 13.87 & 99.94 & 19.99 & 5.29  & 15.82 & 8.53  & 41.89 \\
    STD   & 3.04  & 2.21  & 6.68  & 3.38  & 5.63  & 37.98 & 7.60  & 1.54  & 4.30  & 3.32  & 23.59 \\
    Median & 8.66  & 4.12  & 11.57 & 7.45  & 12.42 & 91.10 & 18.22 & 5.14  & 16.44 & 6.60  & 34.57 \\
    \bottomrule
    \end{tabular}%
  \label{tab:wristscores}%
\end{table*}

Spearman’s correlation coefficients and corresponding p-values are depicted in Tables~\ref{tab:wristrho} and \ref{tab:wristpval}. Most notably, MoCA score and door completion time were negatively correlated (r~=~-0.78, p~=~0.004), and MEp and JNDv were positively correlated (r~=~0.72, p~=~0.013). A similar correlation analysis with Weber fraction conversion did not change significance. Converting handedness scores to absolute values to reflect a degree of ambidexterity, or lack thereof, also did not reveal any significant correlations.

\begin{table*}[tb]
  \centering
  \caption{Spearman’s linear correlation coefficients between all unique comparisons of clinical and robotic test measures across all participants.}
    \begin{tabular}{l|rrrrrrrrr}
    $\rho$   & \multicolumn{1}{r}{MoCA} & \multicolumn{1}{r}{MEg} & \multicolumn{1}{r}{JNDp} & \multicolumn{1}{r}{JNDv} & \multicolumn{1}{r}{JNDt} & \multicolumn{1}{r}{MEp} & \multicolumn{1}{r}{MEv} & \multicolumn{1}{r}{Tk} & \multicolumn{1}{r}{Td} \\
    \midrule
    Handedness & 0.31  & 0.18  & -0.43 & 0.15  & -0.41 & 0.45  & -0.24 & -0.40 & -0.57 \\
    MoCA  &       & -0.33 & -0.47 & -0.17 & -0.28 & 0.14  & -0.19 & -0.29 & -0.78 \\
    MEg   &       &       & -0.02 & 0.38  & -0.33 & -0.02 & -0.17 & 0.37  & 0.06 \\
    JNDp  &       &       &       & 0.06  & 0.07  & 0.25  & 0.14  & 0.30  & 0.59 \\
    JNDv  &       &       &       &       & 0.14  & 0.72  & 0.40  & 0.10  & -0.03 \\
    JNDt  &       &       &       &       &       & -0.12 & 0.09  & 0.05  & 0.29 \\
    MEp   &       &       &       &       &       &       & 0.30  & -0.41 & -0.35 \\
    MEv   &       &       &       &       &       &       &       & 0.58  & 0.34 \\
    Tk    &       &       &       &       &       &       &       &       & 0.39 \\
    \bottomrule
    \end{tabular}%
  \label{tab:wristrho}%
\end{table*}%

\begin{table*}[tb]
  \centering
  \caption{Corresponding p-values to Spearman’s linear correlation coefficients between test measures.}
    \begin{tabular}{l|rrrrrrrrr}
    p-value & \multicolumn{1}{r}{MoCA} & \multicolumn{1}{r}{MEg} & \multicolumn{1}{r}{JNDp} & \multicolumn{1}{r}{JNDv} & \multicolumn{1}{r}{JNDt} & \multicolumn{1}{r}{MEp} & \multicolumn{1}{r}{MEv} & \multicolumn{1}{r}{Tk} & \multicolumn{1}{r}{Td} \\
    \midrule
    Handedness & 0.346 & 0.589 & 0.185 & 0.652 & 0.211 & 0.169 & 0.472 & 0.221 & 0.064 \\
    MoCA  &       & 0.318 & 0.142 & 0.626 & 0.398 & 0.673 & 0.566 & 0.382 & 0.004 \\
    MEg   &       &       & 0.959 & 0.247 & 0.322 & 0.959 & 0.617 & 0.261 & 0.860 \\
    JNDp  &       &       &       & 0.862 & 0.833 & 0.457 & 0.680 & 0.371 & 0.061 \\
    JNDv  &       &       &       &       & 0.686 & 0.013 & 0.225 & 0.776 & 0.946 \\
    JNDt  &       &       &       &       &       & 0.735 & 0.788 & 0.884 & 0.384 \\
    MEp   &       &       &       &       &       &       & 0.364 & 0.214 & 0.299 \\
    MEv   &       &       &       &       &       &       &       & 0.066 & 0.313 \\
    Tk    &       &       &       &       &       &       &       &       & 0.237 \\
    \bottomrule
    \end{tabular}%
  \label{tab:wristpval}%
\end{table*}%

\subsection{Velocity Measures Differ Between Active and Passive, but Position Measures are Inconclusive}

A tabular comparison of position sense measures, as well as velocity sense measures, are listed in Table~\ref{tab:wristpairwise}. From a within-subjects analysis of variance, differences in expected scores of velocity discrimination and matching were statistically significant (p~=~0.000976). Gauge matching, where position was passively sensed and actively pointed contralaterally, and ipsilateral position discrimination also differed significantly in expected score (p~=~0.019). However, other comparisons of position sense, especially between ipsilateral discrimination (passive) and matching (active), were not statistically different (p$>$0.05).

\begin{table}[tb]
  \centering
  \caption{P-values from Friedman’s test to determine any significantly different position test score distributions, and the p-value from Wilcoxon signed-rank test to compare velocity test score distributions between passive and active senses.}
    \begin{tabular}{llr}
    \toprule
    Test 1 & Test 2 & \multicolumn{1}{r}{p-value} \\
    \midrule
    MEg   & JNDp  & 0.019 \\
    MEg   & MEp   & 0.057 \\
    JNDp  & MEp   & 0.859 \\
    \midrule
    JNDv  & MEv   & 0.000976 \\
    \bottomrule
    \end{tabular}%
  \label{tab:wristpairwise}%
\end{table}%

\subsection{Discrimination Tests and Simulated Door Opening are Generally More Demanding than Other Tasks}

\begin{figure}[tb] 
   \centering
   \includegraphics[width=0.95\columnwidth]{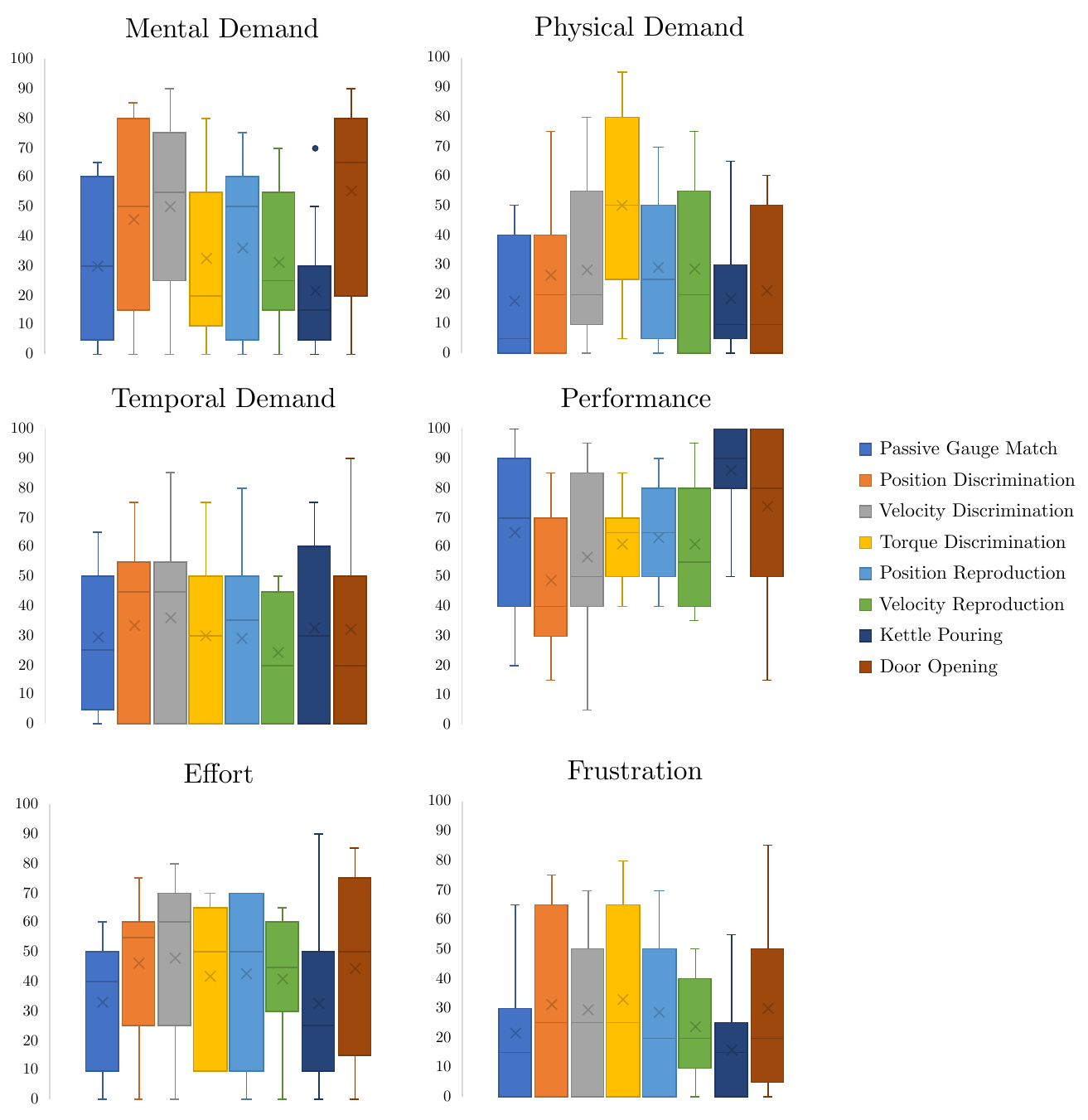}
   \caption{User subjective responses to different questions about task comfort, ease, and demand, aggregated into different test types by color. Scores are aggregated across individuals, with the mid-box lines as sample medians, and vertical x-mark positions indicating sample means.}
   \label{fig:wristTLX}
\end{figure}

Box-and-whisker charts of NASA-TLX response distributions from the post-task surveys are shown in Figure~\ref{fig:wristTLX}. It should be noted that mental demand tended to be higher in position and velocity discrimination, as well as door opening. Torque discrimination had higher physical demand than other tasks. Furthermore, door opening and torque discrimination had a wide range of perceived effort and frustration reported across participants.

\section{Discussion}
The primary goal of this WRIST testbed in its current stage was to investigate proprioceptive acuity and functional ability at the human wrist, and identify distinctive trends of association between sensorimotor modalities that can eventually point to specific types of deficit. As a sanity check for implementation, our average values of positional JND, torque JND, and positional ME generally agree with prior literature \cite{Cappello2015WristAssessment,Contu2018WristPatients,Vicentini2010EvaluationSystem} within an order of magnitude.

Generally, correlation analysis could not find monotonic ranking associations between psychometric measures except for the positive correlation between velocity JND and position ME. This particular association between measures complicates the notion of separate kinesthetic senses, but means our first prediction of demonstrating at least one rank correlation between test measures has been proven.  Perhaps, being harder to tell between pronation velocities may influence potential overshoot in actively reproducing a specific angle from rest. A related explanation could be that senses of position and velocity in healthy humans may share neural pathways depending on task properties. If a patient population also exhibited a similar trend in ability between both velocity and position senses, it may be possible to compensate for any related deficit using other sensory aspects that are more disassociated like vision. The broader observation of null results and weaker correlations between measures otherwise does not reveal any redundant aspects of multimodal integration yet at this stage of investigation. This might imply that there are numerous separate senses and contexts that are integrated, and that more time and effort through additional tests are needed to fully characterize the sensory ``fingerprint'' of a person, healthy or impaired. A partial sensory ``fingerprint'' does not need all tests conducted on the person, but there is a likelihood its ability to identify, diagnose, and treat would be limited. However, these null results may change into significant correlations with further sampling of healthy subjects or new sampling of patients, conducting test-retest sessions, refining or replacing assessments, testing other potential aspects like effort or heaviness, or testing other anatomical axes of the wrist and upper limb.

A positive correlation between task duration of door opening and MoCA score could be due to cognitive effort needed. More specifically, pressing a series of contralateral buttons in a particularly long sequence may involve substantial cognitive memory, abstract thinking, bimanual coordination, and attention. However, the high near-saturation of MoCA scores from healthy participants means that results from this clinical assessment during this study is not compelling evidence. High mental demand reported for the door task compared to other tasks may offer better support instead. Surprisingly though, reported frustration for the door task was low, so virtual door opening with contralateral button mapping was not so hard to cause participants to give up. The role of difficulty in this simulated task can be investigated in future studies by modifying or replacing task sub-components. For example, the interface can be simplified by making the buttons more intuitive, or only require one hand. Door interaction could be changed so that object manipulation only needs user grip and torque on a fixed interface without additional button input. Visual display could be disabled at certain points, and timers of variable length could be added in sequence steps to increase temporal pressure. Finally, physical properties (i.e.,~stiffness, mass) of the moving objects can be set to different levels. Kinematic/kinetic analysis on user input (e.g.,~smoothness) of different difficulty levels could also tease apart features of frustration and diminished motor execution during the task.

In contrast, the kettle opening task reported much less mental demand compared to other tasks. This virtual ADL task is much shorter on average and uses only one hand, so it does not need as much memory or abstraction. Both ADL tasks felt better in perceived success for participants than the psychometric tasks, likely due to the incorporation of visual feedback from ADL completion. Again, modulating task difficulty with sub-components of this kettle task in future studies may tease apart the role of perceived task load. Variables such as increased kettle weight and fluid viscosity, quicker time limits, restricted vision, stricter limits on spilling outside the cup, and distance between the cup and kettle can all determine kettle pour difficulty at different levels.

It is possible to expand beyond rank correlation analysis to use sensory fingerprints to guide functional improvement. The ADL simulations can reflect generalizable functional ability that a patient desires to recover. Meanwhile, the sensory assessments may reflect underlying components not directly observable by the patient during everyday interaction, though these hidden components may still influence recovery in important ways. It is possible to find rank associations and factor interactions between high psychometric scores and high ADL scores in the future. If ADL performance is informed from certain patterns of multimodal integration, it should be possible to use sensory fingerprints to predict how well a person will do in virtual ADLs. Therefore, multivariate multiple regression can use the scores of different psychometric acuity measures as predictors and the performance scores of different ADL simulations as responses. If sensory acuity and integration can be improved through discrimination training distinct from assessment, then it should follow that their functional ability in ADL simulations will improve. Hopefully, this functional improvement would extend beyond specific compensation techniques while training for a particular ADL or assessment. Ultimately, we want to advance from simple patterns of ranking associations to meaningful predictive training in future research.

Based on the robotic psychometric tasks, there is a conditional separation between proprioceptive sensations, especially between active and passive position senses, between velocity and position senses, and internal and external position senses. Our third hypothesis that velocity measures are distinct was demonstrated by pairwise comparison. From significantly different measures of JND and ME for velocity, it seems that motion rate is perceived differently depending on whether or not muscles are being activated. Given that contractile units run parallel to spindles sensitive to rate of changing length, there could be induced lack of tension that may make active velocity sensing less precise than passive sensing. Meanwhile, only passive gauge matching and passive discrimination significantly differed in measure for position sensing. This finding is consistent with the theory that internal relative matching of a limb utilizes different neural pathways than external pointing that requires a body schema to the extrapersonal environment. This study could not establish that passive and active position senses at the wrist are dissociable, even with the lack of statistically significant correlation and redundancy. Therefore, we can only prove our second hypothesis of distinct position senses conditionally. However, this conclusion may change with larger sample sizing or different reference positions.

There are some important caveats to point out about this initial research implementing the new WRIST Testbed. Despite having multiple participants doing multiple task repetitions, the temporal reliability of scores between separate test sessions was not tested due to a limited research time window. Establishing test-retest reliability and evidence for reliable ``fingerprints'' would be necessary to make the tools developed in this study useful for long-term rehabilitation purposes. Another limitation is that contralateral gauge matching in its current implementation, despite its established scientific and clinical use, has significant hurdles in generalizability. To replicate the robot-moved wrist angle, the user has to move their whole opposite arm to adjust a visual indicator placed perpendicular to the plane of robotic rotation. There are several confounding variables that may impact accurate reproduction, including rotated frames of references between hands, activating a whole arm to represent passive rotation of another wrist, the option to verbally confirm their wrist angle afterwards, and the necessity of multiple modalities like vision to report. One might be great at internally determining their robot-moved wrist is at a certain angle, but another deficit may interfere with their ability to reproduce that angle with the other hand and vision. Measuring contralateral matching would be improved in the future by mirroring with the contralateral wrist in the same reference frame of pronation, which would avoid additional steps of cognitive processing and does not necessarily need vision. However, additional hardware would have to be added to the testbed, and test aspects like interrater and test-retest reliabilities needs to be characterized.

Regarding future testbed improvements in implementation, a better cylindrical grip interface may be possible for eventual use in investigating stroke-affected populations. More kinematic measures like rotational smoothness and arm impedance from grip effort can be computed and compared to reveal other potential correlations and differences. We also would like to implement a better velocity estimator to improve the robustness of simulating very stiff virtual walls, which would be helpful to test human ability to reproduce perceived torques. If this device is to be used in a clinical setting, a test-retest procedure should be conducted with certain participants over multiple sessions across days. To generalize functional ability more, we can test out different kinds of ADL simulations such as ball balancing \cite{Yeh2021EffectsSurvivors}. Furthermore, the high physical demand reported by participants after torque discrimination may be due to difficulty in maintaining a consistent motor command from repeating a sustained isometric task. Fatigue can impact a user's ability to produce force \cite{Fisher2017AcuteExercise}, and intensity of voluntary isometric control can impact perceived physical effort \cite{Burgess1995TheControl}. Our finding regarding physical demand in torque discrimination suggests that a lighter reference torque or a quicker ramp-and-return stimulus is needed in the future. 

Our plan is to extend this protocol toward participants affected by stroke or other neurological conditions to investigate the impact of disease on kinesthetic senses better. Furthermore, we want to utilize the testbed’s modularity to test abduction and flexion to validate or reveal additional normative values of kinesthetic acuity. As datasets get exceptionally large for researchers to manually handle, machine learning techniques may be helpful in continuing regression and correlation analyses to make more robust sensory fingerprint profiles. By building up reference datasets for all directions and distinct aspects of sensory perception for more neurological conditions, we can identify each person’s unique ``sensory fingerprint'' that can inform specific and quick treatment for faster neurological recovery.

\section*{Acknowledgment}
We thank John W Krakauer for his insight and help on the project.

This research was developed with funding from the Sloan Foundation and the Interdisciplinary
Rehabilitation Engineering Research Career Development Program.


\bibliographystyle{IEEEtran}
\bibliography{references.bib}

\end{document}